\documentclass[10pt]{article}
\usepackage[margin=1in]{geometry}
\usepackage[utf8]{inputenc}
\usepackage[T1]{fontenc}
\usepackage{lmodern}
\usepackage{textcomp}
\usepackage{amsmath,amssymb}
\usepackage{graphicx,booktabs,longtable,array,calc,microtype,url}
\usepackage{algorithm}
\usepackage{algpseudocode}
\usepackage[hidelinks]{hyperref}
\providecommand{\tightlist}{\setlength{\itemsep}{0pt}\setlength{\parskip}{0pt}}
\newcommand{\pandocbounded}[1]{\resizebox{\linewidth}{!}{#1}}
\title{Phase-Decoupled, Model-Calibrated Power Control\\ for Disaggregated LLM Serving}
\author{Jae Gon Kim, Donghoon Yoo, Hanyul Ryu, Sungho Ha, Juyeon Lee, Soojung Ryu\\[4pt]\normalsize Xenoscube, Inc.\\ \small \url{https://xenoscube.ai/}}
\date{\small September 10, 2026}

\begin{document}
\maketitle

\begin{abstract}
Datacenter GPU power is the binding constraint on LLM serving capacity,
and production serving has shifted to prefill/decode (PD)
disaggregation. Deploying NVIDIA's Max-Q inference profile in-flight on
a disaggregated B200 system, we found its realized gain modest (+8.6\%
tokens/J), model-dependent, and carrying a mean end-to-end latency cost
(+5.2\%) that throughput-only evaluation does not surface; the profile
also applies one setting to prefill and decode GPUs that operate in
opposite hardware regimes. We hypothesize that the optimal power setting
is a property of the deployed (model, quantization, engine, hardware)
combination rather than of the GPU class, that each lane warrants its
own profile, and that converting SLO headroom into energy safely
requires latency-gated calibration under a runtime SLO guard rather than
a fixed recipe. We present a phase-decoupled, model-calibrated
controller: the prefill lane runs under an SM-clock window whose floor
is a latency guarantee by construction, and the decode lane under a
power cap placed by automatic calibration just above a measured
throughput/latency cliff. Because a disaggregated decode lane draws
flat, memory-bound power, the cap binds continuously, the
reactive-overshoot weakness that led POLCA to reject capping is absent,
and the GPU's own power manager retains throughput under the cap. On an
8\ensuremath{\times} B200 node serving Qwen3-Coder-480B (FP8) under
agentic load, our balanced mode delivers +20.4\% tokens/J at +3.5\% mean
e2e versus +8.6\% at +5.2\% for Max-Q, a Pareto improvement on both
axes. On Qwen3-235B-A22B (NVFP4) every operating mode meets the ITL-p99
SLO in every repetition, where both vendor profiles miss it in some. A
decode-actuator A/B shows the calibrated cap beats both
throughput-optimal and aggressive static clock locks (+24.1\% vs +17.2\%
vs +15.5\% tokens/J), and a three-day sustained run saves 32.3\% of a
lane pair's electricity. Both models are Mixture-of-Experts (MoE); an
exploratory dense-model comparison recovers roughly 5\ensuremath{\times}
less, so we scope our claims to MoE serving.

\end{abstract}

\section{Introduction}\label{introduction}

Power, not silicon, is the marginal resource of LLM serving: GPU clusters saturate datacenter power envelopes before they exhaust rack space, so every watt shaved from serving becomes deployable
capacity \cite{polca,splitwise}. Three recipe families dominate practice.

\begin{enumerate}
\def\labelenumi{\arabic{enumi}.}
\tightlist
\item
  \textbf{Vendor static profiles.} NVIDIA's workload power profiles (``Max-Q Inference'') apply a fixed, per-GPU-class multi-knob recipe (TGP limit, Fmax cap, DVFS curve, memory clock); for LLM
  inference the reported gains are 9-12\% datacenter power savings at 2-3\% performance loss \cite{maxq}. The recipe is \emph{phase-blind}: the same setting governs a GPU doing compute-bound prefill
  and one doing memory-bound decode, and it is validated per GPU class, not per model or serving stack.
\item
  \textbf{Cluster power-management frameworks.} POLCA \cite{polca} characterized LLM power on A100s and concluded that \emph{frequency locking} is the reliable reclaim mechanism, explicitly rejecting
  \emph{power capping} as reactive and unpredictable. TAPAS \cite{tapas} manages thermal and power at cluster scale through placement and routing.
\item
  \textbf{Phase-aware DVFS controllers (2025-26).} VoltanaLLM \cite{voltana}, DualScale \cite{dualscale}, GreenLLM \cite{greenllm}, throttLL'eM \cite{throttllem} and DynamoLLM \cite{dynamollm}
  recognize the prefill/decode asymmetry and control energy per phase. Without exception their actuator is frequency selection, and a 2026 characterization study argues that decode power capping is an
  \emph{illusion} that clock locking dominates \cite{illusion}.
\end{enumerate}

In parallel, production stacks (Splitwise \cite{splitwise}, DistServe \cite{distserve}, Mooncake \cite{mooncake}, NVIDIA Dynamo) disaggregate prefill and decode onto separate GPU pools because the two
phases have opposite bottlenecks: prefill is compute-bound and power-hungry; decode is memory-bandwidth-bound and draws flat, stable power well below TDP. Disaggregation turns the transient phase
distinction that prior power work had to chase in time into a static spatial property of a GPU lane. The phase-aware DVFS line exploits this for \emph{when} to actuate, but inherited \emph{what} to
actuate (clocks) from the POLCA-era consensus without re-examining it under disaggregation.

\textbf{Motivating observation.} Deploying recipe (1) on our disaggregated B200 system yielded +8.6\% tokens/J at +5.2\% mean end-to-end latency on a 480B agentic workload (\S{}5.1): a genuine but
modest improvement, with a latency cost that the profile's throughput-oriented validation does not surface. At the same time, even at full target concurrency, the deployment served every request with
visible headroom against its latency SLO: the GPUs ran as fast as the silicon allows to finish work the service contract never asked to be finished that fast. Can that headroom be converted into
energy rather than burned as unrequested speed (\S{}5.2), and can a controller that derives its settings from the deployed stack convert more of it than one fixed, conservative profile (\S{}5.1)? None
of the three recipe families derives its settings from the specific combination being served, so none can locate where the convertible headroom lies. We hypothesize:

\begin{itemize}
\tightlist
\item
  \textbf{H1 (per-combination profile).} The optimal power setting is a property of the deployed (model, quantization, engine stack, hardware) combination, not of the GPU class. It must therefore be
  \emph{measured per combination}, which is practical only if calibration is automatic.
\item
  \textbf{H2 (per-phase profile).} Under PD-disaggregation, a \emph{distinct} profile for the prefill lane and the decode lane recovers energy that any single phase-blind setting necessarily forgoes,
  since no single operating point suits two opposing hardware regimes.
\item
  \textbf{H3 (SLO-gated control).} Converting the headroom safely requires setpoints accepted against the tail SLO and a runtime guard that enforces it when conditions shift. A static class profile
  can do neither, so it must either leave margin it cannot spend or violate the tail when conditions it cannot react to arrive.
\end{itemize}

Our experiments support H1 and H2 directly, and H3 in part (\S{}5): our calibrated operating modes achieve 1.5-3.4\ensuremath{\times} the vendor profile's efficiency gain, and every latency-gated mode
holds the tail SLO in every repetition where both static vendor profiles miss it in a fraction of runs (\S{}5.1). We are explicit about what the data does not isolate: the runtime guard never had
cause to fire in any recorded run, so the compliance contrast is attributable to latency-gated calibration rather than to runtime intervention, and a direct test of the guard remains outstanding
(\S{}6). The design principle throughout is that \emph{each element of a power-control policy (the mechanism, its setpoint, and the metric used to validate it) should be derived from measurements of
the deployed stack}; at several points conclusions drawn from intuition or from throughput-only evaluation were contradicted by measurement, including our own initial actuator assignment (\S{}3.1).

\textbf{A finding beyond the hypotheses.} Testing H2 showed that the two lanes benefit from different \emph{mechanisms}, not merely different settings. On a disaggregated decode lane, a calibrated
power cap outperforms static clock locking, the actuator POLCA recommended: calibrated below the lane's natural draw, the cap binds continuously (contrary to the characterization in \cite{illusion},
such a cap is not inert), and the GPU's built-in power manager apportions frequency under it to preserve throughput, whereas a low lock collapses throughput (-31\% to -33\% in our A/B) and a high lock
saves little (\S{}5.3). Prefill, being compute-bound, is the lane where clocks belong: power tracks SM clock tightly, and a calibrated clock \emph{window} whose floor is chosen from measurement gives
a latency guarantee by construction. Both setpoints must be calibrated per (model, quantization, engine stack): the decode cliff and the natural draw move with engine configuration (CUDA-graph
settings, KV-transfer backend, engine version) as well as with model and quantization, and a roughly one-hour automated pass tracks this while recording the engine fingerprint as a first-class
artifact. Our A/B compares the cap against \emph{static} locks; a comparison against adaptive per-iteration DVFS controllers \cite{voltana,dualscale} is outside this paper's evidence (\S{}6).

\textbf{Contributions.}

\begin{enumerate}
\def\labelenumi{\arabic{enumi}.}
\tightlist
\item
  \textbf{System (\S{}3).} A phase-decoupled actuation design with \emph{heterogeneous mechanisms per lane} (prefill: DVFS window ladder; decode: calibrated power cap), fully automatic per-(model,
  quantization, engine) calibration with a latency-gated acceptance criterion, and an operating-mode ladder (PERF/BAL/EFF) that exposes the efficiency/latency trade as an operator-selectable,
  SLO-guarded knob. Phase-aware energy control for disaggregated serving is not new \cite{voltana,dualscale}; choosing the mechanism per lane from measurement, calibrating it per serving stack, and
  gating it on tail latency are.
\item
  \textbf{Measurement (\S{}5).} The first head-to-head, in-flight comparison against the shipped NVIDIA Max-Q profiles on B200 disaggregated serving, across two MoE models (480B FP8, 235B NVFP4) and
  two workload mixes, extended along the load axis (nine concurrency points) and the time axis (three 24-hour runs). Our modes Pareto-dominate Max-Q on the agentic workload and contain its operating
  point inside our ladder on the standard workload.
\item
  \textbf{Methodology (\S{}4, \S{}5.4).} Evidence that throughput-only metrics hide end-to-end latency damage; a mean-e2e (Little's law) + ITL-p99 evaluation lens; and a repeated-calibration study
  showing that cliff estimation on decode's flat throughput-power curve is noise-fragile and that throughput-only acceptance admits SLO-violating setpoints.
\end{enumerate}

\textbf{Scope.} We control steady-state serving efficiency on a single node's lanes, on MoE models. The mechanism we exploit, decode-lane power slack, is largest where routing sparsity leaves the lane
under-saturated; Appendix A shows the same cap-based mechanism recovers roughly 5\ensuremath{\times} less efficiency on a dense model of comparable scale, so we make no dense-model claims. POLCA's
cluster-level oversubscription goal is complementary (\S{}6). Training is out of scope: disaggregation's spatial phase separation does not exist there, and Zeus \cite{zeus}, Perseus \cite{perseus} and
EnvPipe \cite{envpipe} already cover training energy.

\section{Background and Motivation}\label{background-and-motivation}

\subsection{PD-disaggregated serving}\label{pd-disaggregated-serving}

An LLM request has two phases with opposite hardware signatures. \emph{Prefill} processes the whole prompt in parallel: high arithmetic intensity, tensor cores saturated, power spiking toward TDP.
\emph{Decode} emits tokens one at a time: dominated by weight and KV-cache reads, memory-bandwidth-bound, with GPU power flat and well under TDP. Serving stacks disaggregate the phases onto separate
GPU pools connected by a KV-cache transfer path (Figure 1, top) to protect decode's inter-token latency (ITL) from prefill interference and to scale the pools independently
\cite{splitwise,distserve,mooncake}.

Disaggregation has an under-exploited corollary for power management: each GPU's bottleneck regime is now known statically. A decode-pool GPU remains memory-bound for the entirety of its service
assignment, so the phase-identification problem that dominated prior phase-aware power work reduces to lane identity.

\subsection{GPU power knobs, and what prior work concluded}\label{gpu-power-knobs-and-what-prior-work-concluded}

GPUs expose two in-band knobs: \emph{frequency locking} (pin the SM clock) and \emph{power capping} (the on-device power manager throttles clocks whenever draw would exceed the cap). POLCA
\cite{polca}, characterizing A100-era colocated serving, observed that capping is \emph{reactive}, since prompt-phase spikes can overshoot the cap before throttling engages, while frequency locking
reclaims power predictably. It built its policy on clock caps while conceding their cost: \emph{``frequency locking incurs performance impact throughout the execution, and not just when the power
utilization is high.''} Under disaggregation both premises flip. The overshoot scenario requires prefill spikes, and a decode lane has none: capping a decode lane is steady-state control of a flat
signal. And the lock's constant tax is no longer acceptable, because on a memory-bound lane the throughput-vs-power curve is flat near the top and cliff-like below; the efficient operating point sits
just above a cliff that a static clock cannot find, and pinning clocks discards the power manager's ability to reallocate budget between SM and memory dynamically.

\textbf{An analytical lens on the mechanism split.} The standard CMOS model puts GPU power at

\begin{equation}
P \;\approx\; P_{\text{static}} + CV^2 f \;\sim\; A + B f^{1+\alpha}, \qquad \alpha > 0
\end{equation}

since voltage co-scales with frequency. Writing execution time as \(T \sim f^{-\beta}\), with \(\beta \in [0,1]\) the workload's arithmetic-intensity coefficient (\(\beta{=}1\) fully compute-bound,
\(\beta{=}0\) fully memory-bound), energy per unit of work becomes

\begin{equation}
E \;=\; P\,T \;\sim\; A f^{-\beta} + B f^{1+\alpha-\beta}
\end{equation}

VoltanaLLM derives this same model to motivate per-phase \emph{frequency selection} \cite{voltana}; we push it one step further, to the choice of \emph{mechanism}. For the prefill lane
\(\beta \approx 1\): time responds nearly proportionally to clock, so a calibrated clock floor is a latency guarantee by construction, and power tracks \(f^{1+\alpha}\), which makes the clock window
the natural efficiency knob. For the decode lane \(\beta \ll 1\): lowering frequency buys power at a sub-linear time cost, but only down to the bandwidth-saturation point below which the pipeline
collapses (the cliff of \S{}3.3) \cite{voltana,dualscale}. Moreover \(\beta\) rises with batch occupancy, so the energy-optimal frequency \emph{moves with load}, and a static lock must sit
conservatively high or risk crossing the moving optimum. The phase-aware DVFS line answers by predicting the optimum in software per iteration; a calibrated cap answers by fixing the power budget and
delegating the frequency decision to the GPU's own power manager, which re-solves the allocation continuously at microsecond scale with no model to mistrain. Our A/B (\S{}5.3) tests the cap against
the static-lock answer; the software-predicted answer is not in our evidence.

NVIDIA's workload power profiles \cite{maxq} take a simpler position: one validated multi-knob recipe per GPU class, applied uniformly. This is deployable in minutes and is the natural production
baseline, which is why we benchmark against the real in-flight profiles (\S{}4). NVIDIA's own evaluation reports that multi-knob profiles \emph{including a TGP power limit} outperform pure frequency
scaling by 7-9\% in performance at equal power savings on B200 \cite{maxq}, vendor-reported evidence of the mechanism asymmetry we isolate per phase.

\subsection{Why calibration must be automatic and engine-aware}\label{why-calibration-must-be-automatic-and-engine-aware}

The decode cliff (the cap below which throughput and latency collapse) and the natural power (the uncapped draw that anchors the prefill window) are functions of model, quantization, \emph{and the
serving engine's configuration}. Across engine configurations of the same model on the same GPUs we measured natural decode power move by roughly 3\%, and observed ITL shifts across engine versions of
the same stack. A 3\% shift is comparable to the run-to-run dispersion of our short-run measurements (\S{}5.1), so we take it as motivation for fingerprinting rather than as a result in itself; the
operational point is that a profile validated on one stack has no mechanism to notice that the next stack differs. Our calibration pass (\S{}3.3) therefore records the engine fingerprint (CUDA-graph
on/off, KV-transfer backend, engine version) alongside the setpoints, and any fingerprint change triggers recalibration.

\section{Design}\label{design}

Figure 1 shows the system. The serving plane is untouched, stock NVIDIA Dynamo + sglang-runtime PD-disaggregated serving. Our control plane attaches per-lane actuators driven by calibration artifacts,
an operating-mode selector, and an SLO guard.

\begin{figure}[tbp]
\centering
\pandocbounded{\includegraphics[keepaspectratio,alt={Phase-decoupled control architecture. The serving plane is an unmodified PD-disaggregated stack; the control plane assigns each lane its own actuator (SM-clock window for prefill, calibrated power cap for decode), parameterized by a per-(model, quant, engine) calibration store, an operating-mode ladder, and an ITL-p99 SLO guard.}]{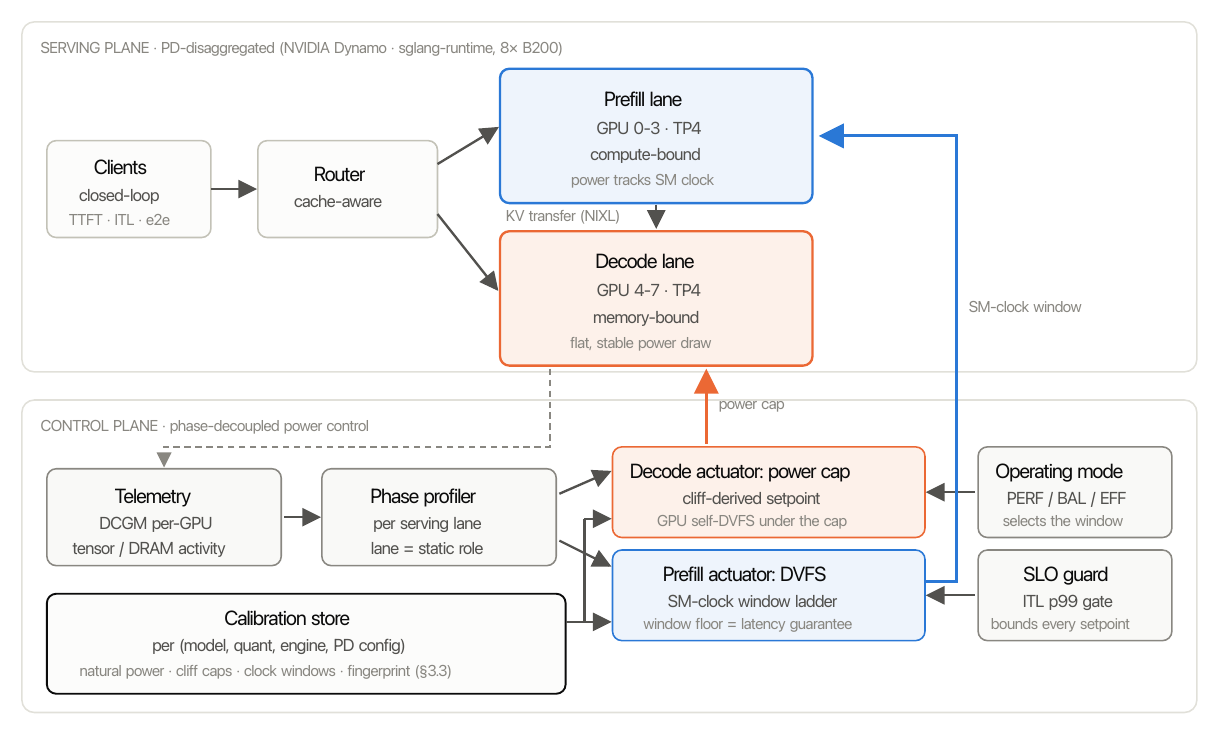}}
\caption{Phase-decoupled control architecture. The serving plane is an unmodified PD-disaggregated stack; the control plane assigns each lane its own actuator (SM-clock window for prefill, calibrated
power cap for decode), parameterized by a per-(model, quant, engine) calibration store, an operating-mode ladder, and an ITL-p99 SLO guard.}
\end{figure}

\textbf{Problem statement.} A PD-disaggregated deployment exposes two lanes \(\ell \in \{P, D\}\) with independent actuators: a prefill SM-clock window \(W = [f_{\text{lo}}, f_{\text{hi}}]\) and a
decode power cap \(c\). Given the service's tail SLO \(\Lambda\) (an ITL-p99 bound, with prompt latency tracked as a secondary guard), the controller solves

\begin{equation}
\max_{(c,\,W)} \;\; \text{tokens/J} \quad \text{s.t.} \quad \text{ITL-p99} \le \Lambda \;\; \text{at the deployment's operating load}
\end{equation}

with \((c, W)\) restricted to the hardware's actuation grid and drawn from a calibration artifact \(\mathcal{A}\) indexed by (model, quantization, engine, hardware). H1 says \(\mathcal{A}\) must be
measured per combination; H2 says it assigns distinct settings, indeed distinct mechanisms, to the two lanes; H3 says the constraint binds at the tail, so feasibility must be established against tail
latency and enforced at runtime. The components are per-lane actuation (\S{}3.1), an operating-mode ladder exposing solutions at three risk postures (\S{}3.2), and the calibration pass that populates
\(\mathcal{A}\) (\S{}3.3).

\subsection{Phase-decoupled actuation}\label{phase-decoupled-actuation}

\textbf{Decode lane: power cap.} The actuator is a single per-GPU power cap chosen from the calibrated cliff (\S{}3.3). The cap bounds peak draw while delegating the SM/memory frequency split to the
GPU's own power manager, which performs microsecond-scale DVFS \emph{under} the cap. This adaptivity is the root cause of capping's win over static clock locks in our A/B (\S{}5.3): the cap achieves
power reduction \emph{and} throughput retention simultaneously, which no single locked frequency can.

\textbf{Prefill lane: SM-clock window ladder.} Prefill power tracks clock nearly proportionally, so we control clocks directly, but as a \emph{window} rather than a point lock. Each operating mode
confines the clock to a calibrated band spanning roughly 20-30\% of the achievable clock range (boundaries are per-combination calibration outputs; the most efficient mode's floor sits near half of
the maximum clock). The window floor is the calibrated clock below which prompt-latency SLOs would be at risk, so selecting a mode selects a latency guarantee by construction. Within the window, a
lightweight runtime controller adjusts the operating clock with lane-level load, its actuation rate bounded so as not to stress the driver stack. The in-window policy belongs to the commercial
implementation behind this section's contracts (see Reproducibility); the guarantees evaluated in \S{}5 derive from the window bounds, and all results are gathered with the controller active. Because
the controller was active in every arm, our measurements do not separate the floor's contribution from the in-window policy's (\S{}5.4, \S{}6).

\textbf{On the assignment of mechanisms to lanes.} The symmetric intuition (``cap the compute-bound phase, DVFS the memory-bound phase'') was our own initial design, and measurement reversed it
(\S{}5.3). Compute-bound prefill responds to clocks predictably enough that a window gives a \emph{guarantee}; memory-bound decode responds to clocks so nonlinearly (flat-then-cliff) that only the
cap's self-DVFS adaptivity extracts the efficiency without collapsing throughput. Both prior recipes (POLCA's all-frequency, Max-Q's uniform profile) encode the un-measured intuition. Appendix A
documents the selection campaign behind this choice, spanning four base models, three quantizations, dense/MoE/hybrid architectures, three workload shapes, and both colocated and disaggregated
topologies.

\subsection{Operating-mode ladder}\label{operating-mode-ladder}

The controller's objective is fixed: maximize energy efficiency subject to the latency SLO. Latency enters as a constraint every setpoint must satisfy, never as a quantity traded in the open.
Operators pick one of three modes per lane pair, each mapping to (decode cap, prefill window) pairs from the calibration artifact; the modes differ only in how much of the calibrated headroom they
spend:

\begin{itemize}
\tightlist
\item
  \textbf{PERF (Performance)} places the decode cap at the lane's natural draw and keeps the prefill window in its upper range. Steady-state behavior is essentially unthrottled; PERF buys a
  \emph{deterministic peak bound} (the cap clips excursions above natural draw) plus a small efficiency gain, at near-baseline latency.
\item
  \textbf{BAL (Balanced)} sets both lanes to intermediate calibrated setpoints. It is the default recommendation: a majority of the EFF-level gain at a substantially smaller latency cost (\S{}5.2).
\item
  \textbf{EFF (Efficiency)} applies the deepest SLO-admissible setpoints the calibration found: the decode cap just above the calibrated cliff and the prefill window floor at the latency guarantee. It
  suits workloads with latency headroom (the agentic mix of \S{}5.1).
\end{itemize}

All three modes come from one calibration artifact, so the ladder is monotone by construction and every rung is subject to the same runtime SLO guard, which monitors tail latency per lane and, on a
violation, releases the offending lane's setpoint toward the hardware default until compliance is restored (Algorithm 1, lines 7-9). This is the axis Max-Q lacks: its single profile is one operating
point, which our measurements place inside the ladder (between PERF and BAL on the standard workload). Algorithm 1 states the runtime loop at contract level; the loop's structure is what a
reimplementation would need to reproduce our SLO-compliance results. Every result in \S{}5 was gathered with the guard armed, except the Figure 7 capture, which states so; in no recorded run did the
guard trip (\S{}5.1, \S{}5.7), so \S{}5 exercises lines 1-6 and 10 of the loop and establishes lines 7-9 by contract rather than by observation.

\begin{algorithm}[t]
\caption{Runtime mode execution with SLO guard (contract level)}
\begin{algorithmic}[1]
\Require calibration artifact $\mathcal{A}$ (\S3.3); operating mode $m \in \{\mathrm{PERF}, \mathrm{BAL}, \mathrm{EFF}\}$; tail SLO $\Lambda$ (ITL-p99); per-lane telemetry stream
\Ensure continuously enforced lane setpoints
\State $(c_m, W_m) \gets \mathcal{A}[m]$ \Comment{decode cap, prefill clock window}
\State apply power cap $c_m$ to every decode-lane GPU
\State confine prefill-lane clocks to $W_m$ \Comment{floor of $W_m$ = latency guarantee}
\Loop\ at each control interval
  \ForAll{lanes $\ell \in \{P, D\}$}
    \State $\lambda_\ell \gets$ observed tail latency \Comment{ITL-p99 (decode), TTFT (prefill)}
    \If{$\lambda_\ell$ violates $\Lambda$}
      \State release lane $\ell$'s setpoint toward the hardware default
      \State hold until compliance is restored, then re-apply $\mathcal{A}[m]$
    \EndIf
  \EndFor
  \State within $W_m$: the runtime controller moves the prefill clock with lane load \par\hskip\algorithmicindent (rate-bounded; policy replaceable --- the \S5 guarantees derive from the window bounds, lines 1--9)
\EndLoop
\end{algorithmic}
\end{algorithm}

\subsection{Automatic calibration}\label{automatic-calibration}

For each (model, quantization, engine) combination, an automated pass runs offline against the live serving stack and produces three outputs: (i) the natural (uncapped) power of each lane, which
anchors all operating windows; (ii) decode power-cap setpoints per mode, each accepted only if it preserves both throughput and the ITL-p99 SLO under serving-parity load; and (iii) prefill
clock-window boundaries per mode. The pass completes in about an hour without operator involvement and records the engine-stack fingerprint alongside the setpoints. The setpoint search is a
replaceable component: any search whose outputs pass the acceptance gate of Algorithm 2 yields valid setpoints; \S{}5.4 characterizes our procedure's output stability, and every setpoint used in the
evaluation is reported in \S{}5.

\begin{figure}[tbp]
\centering
\pandocbounded{\includegraphics[keepaspectratio,alt={Calibration contract and artifact semantics. (a) The setpoint search is replaceable; its contract is fixed: it runs against the live stack, accepts a setpoint only if throughput is preserved and ITL-p99 stays within the SLO, and emits an artifact that a fingerprint change invalidates. (b) What the artifact encodes on each lane's actuation axis: on the decode power axis, the region below the calibrated cliff is rejected by the acceptance gate and the three mode caps span the calibrated headroom between the cliff (EFF) and the natural draw (PERF); on the prefill clock axis, each mode selects a window whose floor is the latency guarantee. Positions are schematic; every marker is a per-combination measurement, not a universal constant.}]{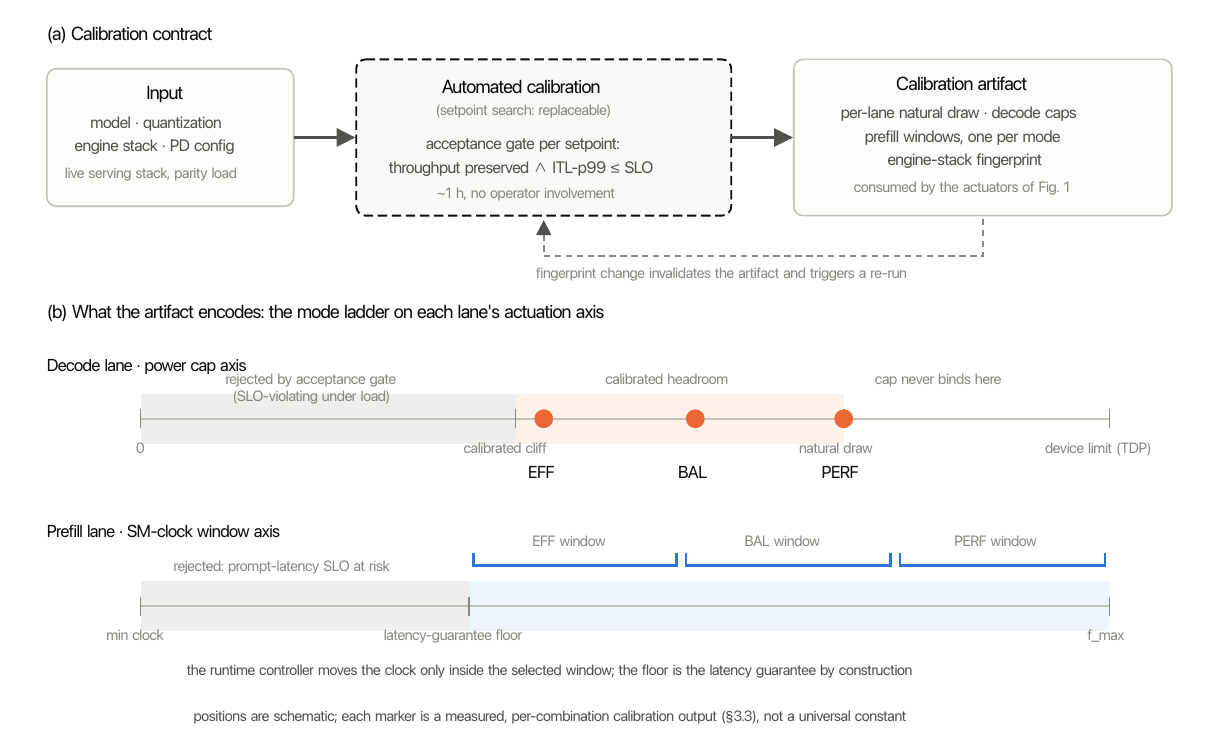}}
\caption{Calibration contract and artifact semantics. (a) The setpoint search is replaceable; its contract is fixed: it runs against the live stack, accepts a setpoint only if throughput is preserved
and ITL-p99 stays within the SLO, and emits an artifact that a fingerprint change invalidates. (b) What the artifact encodes on each lane's actuation axis: on the decode power axis, the region below
the calibrated cliff is rejected by the acceptance gate and the three mode caps span the calibrated headroom between the cliff (EFF) and the natural draw (PERF); on the prefill clock axis, each mode
selects a window whose floor is the latency guarantee. Positions are schematic; every marker is a per-combination measurement, not a universal constant.}
\end{figure}

The relative setpoints reported in \S{}5 (decode caps at 80-90\% of natural draw) are outputs of this calibration for the specific model, quantization, engine stack and PD configuration measured, not
universal constants. They differ between our own two models (\ensuremath{\approx}85\% of natural on 480B TP4 PD versus 80-90\% on 235B TP1 PD), and per H1 no fixed fraction should be expected to
transfer.

\textbf{Latency-aware acceptance is not optional.} An earlier calibrator revision accepted setpoints on throughput alone; on 235B it selected an EFF cap near 65\% of natural draw that preserved
throughput within tolerance while pushing ITL-p99 beyond the SLO in the large majority of repeated runs. Adding a tail-latency criterion moved the accepted setpoint to 80\% of natural draw and
restored 100\% SLO compliance at +22.4\% tokens/J (\S{}5.2). Algorithm 2 fixes this acceptance gate, the part of calibration our correctness claims depend on and that a reader can port to their own
calibrator.

\begin{algorithm}[t]
\caption{Calibration acceptance gate (contract level)}
\begin{algorithmic}[1]
\Require candidate setpoint $u$ (a decode cap or a prefill window floor); tail SLO $\Lambda$; tolerance $\varepsilon$; serving-parity load $L$ (closed loop, same $N$ as service, mixed prefill+decode traffic)
\Ensure accept/reject; on accept, an entry in artifact $\mathcal{A}$
\State $T_{\mathrm{ref}} \gets$ reference throughput at hardware defaults under $L$ \Comment{multi-repetition, noise-robust statistic; a single-run $T_{\mathrm{ref}}$ is unsafe on decode's flat curve (\S5.4)}
\State run the live stack at $u$ under $L$
\State \textbf{accept} $u$ \textbf{iff} $\mathrm{throughput}(u) \ge (1-\varepsilon)\,T_{\mathrm{ref}}$ \textbf{and} $\text{ITL-p99}(u) \le \Lambda$ \Comment{the latency-aware criterion; omitting it admits SLO-violating setpoints (\S5.4)}
\State on accept: record $u$ in $\mathcal{A}$ together with the engine-stack fingerprint (CUDA-graph configuration, KV-transfer backend, engine version)
\State any fingerprint change invalidates $\mathcal{A}$ and triggers recalibration
\end{algorithmic}
\end{algorithm}

\begin{figure}[tbp]
\centering
\pandocbounded{\includegraphics[keepaspectratio,alt={Measured decode cap-sweep from the calibration pass (235B NVFP4, TP1 PD, calibration load). Throughput (left axis) is flat from the natural draw down to the calibrated cliff (430 W, the minimum-J/tok point) and collapses below it (-36\% at 350 W); ITL p95 (right axis) stays well inside the SLO everywhere above the cliff. The calibrated mode caps span the headroom: EFF 460 W just above the cliff, BAL 547 W, PERF uncapped at the natural draw. This is the measured counterpart of Figure 2b's schematic; the sweep reports ITL p95, and the acceptance gate additionally checks ITL-p99 under serving-parity load (Algorithm 2).}]{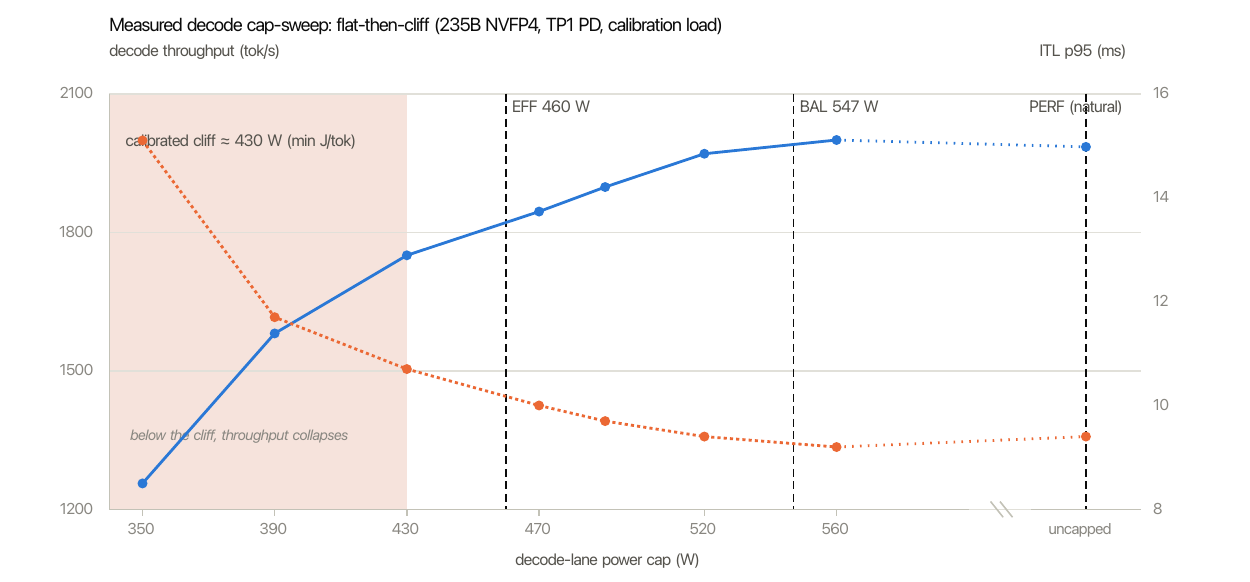}}
\caption{Measured decode cap-sweep from the calibration pass (235B NVFP4, TP1 PD, calibration load). Throughput (left axis) is flat from the natural draw down to the calibrated cliff
(\ensuremath{\approx}430 W, the minimum-J/tok point) and collapses below it (-36\% at 350 W); ITL p95 (right axis) stays well inside the SLO everywhere above the cliff. The calibrated mode caps span
the headroom: EFF 460 W just above the cliff, BAL 547 W, PERF uncapped at the natural draw. This is the measured counterpart of Figure 2b's schematic; the sweep reports ITL p95, and the acceptance
gate additionally checks ITL-p99 under serving-parity load (Algorithm 2).}
\end{figure}

\section{Experimental Setup}\label{experimental-setup}

{\begin{longtable}[]{@{}
  >{\raggedright\arraybackslash}p{(\linewidth - 2\tabcolsep) * \real{0.14}}
  >{\raggedright\arraybackslash}p{(\linewidth - 2\tabcolsep) * \real{0.86}}@{}}
\toprule\noalign{}
\begin{minipage}[b]{\linewidth}\raggedright
Item
\end{minipage} & \begin{minipage}[b]{\linewidth}\raggedright
Value
\end{minipage} \\
\midrule\noalign{}
\endhead
\bottomrule\noalign{}
\endlastfoot
Node & 8\ensuremath{\times} NVIDIA B200 (single node) \\
Serving stack & NVIDIA Dynamo \texttt{sglang-runtime:1.2.0}, PD-disaggregated, cache-aware router \\
Models & Qwen3-Coder-480B FP8 (TP4 prefill GPUs 0-3 / TP4 decode GPUs 4-7); Qwen3-235B-A22B NVFP4 (TP1 PD, GPUs 0-1) \\
Load & Closed-loop at the router; concurrency 48 (480B) / 24 (235B) \\
Workloads & agentic (long-context, tool-call-shaped mix) and standard (chat-shaped mix) \\
Arms & BL (uncontrolled), NVIDIA Max-P (profile 0,6), Max-Q (profile 1,6), both real in-flight profiles rather than reconstructions, and our PERF/BAL/EFF operating modes \\
Campaigns & The head-to-head and ladder results (\S{}5.1-\S{}5.5) are single-load-point studies at the concurrency above. Two further campaigns extend the evaluation along one axis each and are
reported with their own scope: the load axis (\S{}5.6, 235B, nine concurrency points) and the time axis (\S{}5.7, 235B, three 24-hour runs). Their absolute values are not interchangeable with the
tables of \S{}5.1, differing in engine queue depth, GPU pair, and energy-accounting scope, each stated in place. \\
Energy metric & tokens/J over all 8 GPUs (including idle/other lanes; the denominator is never restricted to actively controlled lanes) \\
Latency metrics & mean end-to-end (Little's-law lens: at fixed closed-loop concurrency, mean-e2e \ensuremath{\propto} 1/completion rate), TTFT p95/p99, ITL p95/p99; every arm is judged against a fixed
ITL-p99 service SLO \\
Repetitions & N = 3 per arm and workload cell in the head-to-head campaigns of \S{}5.1 (both models); dispersion is reported as \ensuremath{\pm}1\ensuremath{\sigma} across repetitions where N
\textgreater{} 2. The load-axis campaign of \S{}5.6 uses N = 3 per cell and the time-axis campaign of \S{}5.7 three full days \\
Data drop rule & Intermittent TTFT blowups (\textgreater15 s; stack-level issue, arm-independent, a small fraction of repetitions) dropped pairwise so all arms see identical repetition sets \\
\end{longtable}
}

\textbf{Workload composition and concurrency.} Both workloads are closed-loop request streams issued against the router at fixed concurrency: N logical clients each hold exactly one request in flight
and issue the next the moment the previous completes, so in-flight requests are pinned at N for the entire run. Both lanes therefore stay at a steady operating point, which makes the lane power
profiles stationary enough to control and to measure; load is identical across arms by construction, since an arm that slows down receives requests more slowly and no arm is penalized by queue
buildup; and with N fixed, Little's law ties mean end-to-end latency to the request completion rate exactly (below). N is chosen per deployment to saturate it (48 on the 8-GPU TP4/TP4 configuration,
24 on the 2-GPU TP1 configuration), and calibration runs under the same construction, so setpoints are measured at the load they will serve. The \emph{agentic} workload models tool-using agent
traffic: long, context-accumulating prompts with comparatively long generations, stressing both large prompt ingests and sustained decode. The \emph{standard} workload models interactive chat traffic
and is derived from the public ShareGPT conversation corpus \cite{sharegpt}: request content and input/output length distributions are sampled from ShareGPT conversations, giving shorter prompts and
generations at higher request turnover and a prefill-heavier duty cycle per token served. The two bracket the phase balance a production mix moves between; \S{}5.2 shows the appropriate operating mode
differs between them, which is itself part of the argument for a mode ladder. The generator is seeded per run (\texttt{-\/-seed\ 1234} throughout), so a repetition replays the same request stream. The
standard workload samples input length from \{64, 128, 256, 512, 1024, 2048\} tokens with weights \{0.18, 0.24, 0.26, 0.18, 0.10, 0.04\} and output length from \{32, 64, 128, 256, 512, 1024\} with
weights \{0.16, 0.22, 0.26, 0.20, 0.12, 0.04\}. The agentic workload issues 4 tool-calling turns per client, base input 512 tokens and 128 generated tokens per turn, a tool result of 256 tokens
appended to the context after each turn, and a think-gap drawn uniformly from 0.5-3 s between turns.

\begin{figure}[tbp]
\centering
\pandocbounded{\includegraphics[keepaspectratio,alt={Request anatomy of the two workloads (schematic; segment lengths not to scale). The agentic workload accumulates context across tool-calling turns, so prefill ingests grow while generations remain long; the standard workload issues short independent requests at high turnover, with content and lengths drawn from ShareGPT . The two bracket the prefill/decode balance a production mix moves between.}]{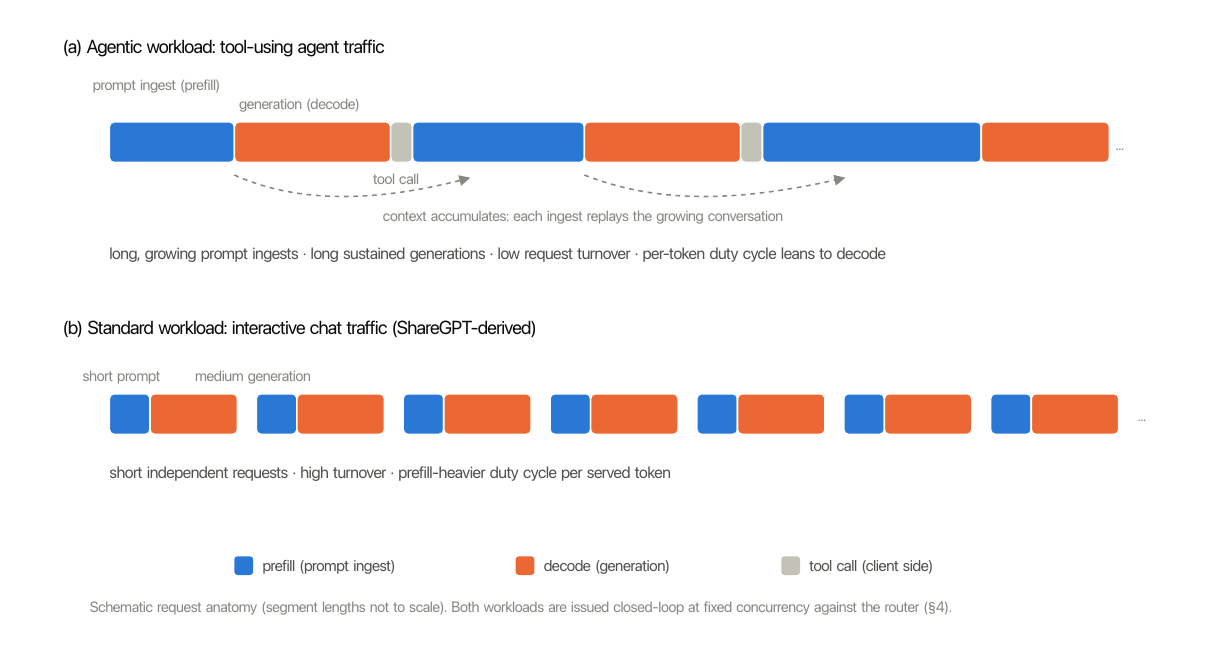}}
\caption{Request anatomy of the two workloads (schematic; segment lengths not to scale). The agentic workload accumulates context across tool-calling turns, so prefill ingests grow while generations
remain long; the standard workload issues short independent requests at high turnover, with content and lengths drawn from ShareGPT \cite{sharegpt}. The two bracket the prefill/decode balance a
production mix moves between.}
\end{figure}

\textbf{Benchmark provenance and arrival control.} The standard workload's requests come from ShareGPT \cite{sharegpt}, the source used by several serving-systems evaluations, so its content and
length distributions are externally anchored. What we deliberately replace is the \emph{arrival process}: both workloads are issued through the closed-loop harness rather than by replaying trace
timestamps, because the paired methodology of \S{}5 requires a stream that is exactly reproducible and load-controlled, and because tokens/J is only a meaningful property of an \emph{operating point}
if load is stationary over the integration window. Native-timestamp replay is open-loop and non-stationary, which breaks the fixed-N construction and injects arrival variance large enough to mask
single-digit-percent effects in energy. The agentic workload is constructed: context accumulating across tool-calling turns dominates our target deployments but is not represented in any public trace.
Evaluating under native trace arrival processes is future work (\S{}6).

\textbf{Why mean-e2e rather than throughput alone.} Throughput-only reporting conceals the latency cost of power control: an arm can maintain aggregate tokens/s while individual requests lengthen. At
fixed closed-loop concurrency \(N\), Little's law ties mean end-to-end latency to the request completion rate \(\mu\) exactly:

\begin{equation}
\overline{t}_{\text{e2e}} \;=\; \frac{N}{\mu}
\end{equation}

Because every arm serves the identical request stream, per-request token counts match across arms and \(\mu\) differs from token goodput only by a constant, so \ensuremath{\Delta}mean-e2e is a
workload-normalized measure of latency degradation with no free parameters; tail ITL (p99) additionally guards the streaming experience. We report both for every arm. Under this lens, settings that
appear cost-free by throughput (including Max-Q's reported results) exhibit a measurable latency cost (\S{}5.1).

\textbf{On the strength of the vendor baseline.} Comparing against a \emph{reconstruction} of the vendor profile rather than the profile itself understates the baseline. Running both in the same block
on the same hardware, a uniform SM-clock lock, the natural reconstruction, saves 4.0-4.1\% of baseline power, while the in-flight profile saves 12.5-12.9\%: a factor of three. Every Max-Q comparison
in this paper uses the shipped profile.

\textbf{Quantization on the actuation grid.} Caps and clocks are quantized (cap grid steps, discrete supported clocks); a policy computed off-grid silently snaps to the grid at actuation time. Our
harness measures at actuated values only.

\textbf{Energy measurement.} Per-GPU power is sampled in-band (NVML, with DCGM activity counters alongside) at a 100 ms cadence; energy is the time integral of measured power over the full run window,
summed across all 8 GPUs of the node. Token counts are taken client-side from the streamed responses, so tokens/J divides externally observed work by internally measured energy and cannot be inflated
by either side alone.

\section{Evaluation}\label{evaluation}

\subsection{Head-to-head vs NVIDIA Max-Q}\label{head-to-head-vs-nvidia-max-q}

\begin{figure}[tbp]
\centering
\pandocbounded{\includegraphics[keepaspectratio,alt={Efficiency/latency plane, agentic workload on both models. (a) 480B: both BAL and EFF lie above and to the left of Max-Q --- 2.4 and 3.4 its efficiency gain at lower mean-e2e --- so each is a Pareto improvement on both axes. (b) 235B (paired repeated runs; error bars 1): the result reproduces on a second model; BAL matches Max-Q's latency cost while exceeding its gain, and EFF delivers 3.4 the gain. Every plotted arm meets the ITL-p99 SLO. The standard-workload ladder, where efficiency is bought with latency inside the SLO envelope, is reported in 5.2.}]{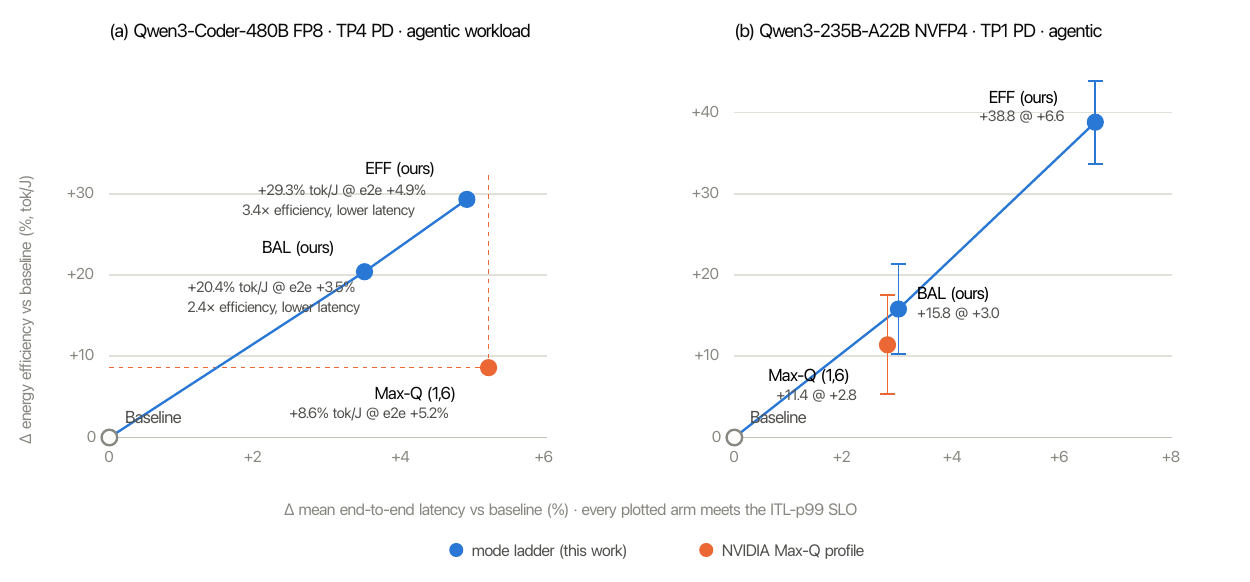}}
\caption{Efficiency/latency plane, agentic workload on both models. (a) 480B: both BAL and EFF lie above and to the left of Max-Q --- 2.4\ensuremath{\times} and 3.4\ensuremath{\times} its efficiency
gain at lower mean-e2e --- so each is a Pareto improvement on both axes. (b) 235B (paired repeated runs; error bars \ensuremath{\pm}1\ensuremath{\sigma}): the result reproduces on a second model; BAL
matches Max-Q's latency cost while exceeding its gain, and EFF delivers 3.4\ensuremath{\times} the gain. Every plotted arm meets the ITL-p99 SLO. The standard-workload ladder, where efficiency is
bought with latency inside the SLO envelope, is reported in \S{}5.2.}
\end{figure}

\textbf{480B agentic (primary result).} In-flight Max-Q yields +8.6\% tokens/J at +5.2\% mean-e2e. Our BAL mode yields +20.4\% at +3.5\% and EFF +29.3\% at +4.9\%: 2.4\ensuremath{\times} and
3.4\ensuremath{\times} the efficiency gain, each at lower latency than the vendor profile, so both are Pareto improvements on both axes. Repeated reconfirmation on the same stack under the standard
workload gives EFF +23.6\ensuremath{\pm}3.7\% and BAL +24.3\ensuremath{\pm}0.7\% tokens/J, with every arm passing the ITL-p99 SLO in every repetition. The 480B agentic figures above are two-repetition
means from a single session (N = 2 per arm), so no \ensuremath{\pm}1\ensuremath{\sigma} is quoted for them; the reconfirmation values that follow carry their own dispersion.

\textbf{235B (repeated paired study).} With calibration frozen and paired repetitions per arm (anomalous repetitions dropped pairwise; Figure 5b):

{\begin{longtable}[]{@{}llllll@{}}
\toprule\noalign{}
Workload & Arm & \ensuremath{\Delta}tok/J & Mean power saved & \ensuremath{\Delta}mean-e2e & ITL-p99 SLO \\
\midrule\noalign{}
\endhead
\bottomrule\noalign{}
\endlastfoot
agentic & Max-P & +0.7\ensuremath{\pm}6.5\% & +0.3\% & -0.1\% & 96\% \\
& Max-Q & +11.4\ensuremath{\pm}6.1\% & +12.4\% & +2.8\% & 96\% \\
& PERF (ours) & +0.2\ensuremath{\pm}5.8\% & +0.4\% & +0.7\% & 100\% \\
& BAL (ours) & +15.8\ensuremath{\pm}5.5\% & +15.8\% & +3.0\% & 100\% \\
& \textbf{EFF (ours)} & \textbf{+38.8\ensuremath{\pm}5.1\%} & \textbf{+32.2\%} & \textbf{+6.6\%} & \textbf{100\%} \\
standard & Max-P & -0.1\ensuremath{\pm}1.7\% & -0.2\% & -0.2\% & 100\% \\
& Max-Q & +11.8\ensuremath{\pm}2.0\% & +12.7\% & +2.8\% & 100\% \\
& PERF (ours) & +7.6\ensuremath{\pm}3.2\% & +10.8\% & +1.7\% & 100\% \\
& \textbf{BAL (ours)} & \textbf{+17.6\ensuremath{\pm}2.6\%} & \textbf{+22.6\%} & \textbf{+3.8\%} & \textbf{100\%} \\
& EFF (ours) & +22.4\ensuremath{\pm}2.4\% & +29.5\% & +8.9\% & 100\% \\
\end{longtable}
}

The three rows per workload marked \emph{(ours)} are the operating-mode ladder of \S{}3.2 in ladder order. Mean power saved is the reduction in mean draw over the same lanes the efficiency figure is
computed on; baseline draw is 799 W on the agentic workload and 896 W on the standard workload. Max-P is statistically indistinguishable from the uncontrolled baseline on both workloads, and on the
agentic workload so is our PERF mode.

Two readings of the table carry the argument. First, on the agentic workload EFF exceeds Max-Q by a margin far outside the \ensuremath{\pm}1\ensuremath{\sigma} bands (+38.8\ensuremath{\pm}5.1\%
against +11.4\ensuremath{\pm}6.1\%) at a +6.6\% mean-e2e cost and a 5.8\% TTFT p95 improvement; BAL's advantage over Max-Q (+15.8\ensuremath{\pm}5.5\% against +11.4\ensuremath{\pm}6.1\%, at a matched
latency cost of +3.0\% versus +2.8\%) lies within the overlap of the bands and we do not claim it as a separate result. On the standard workload, where dispersion is smaller, BAL exceeds Max-Q outside
the bands (+17.6\ensuremath{\pm}2.6\% against +11.8\ensuremath{\pm}2.0\%) at a mean-e2e cost one point above the vendor profile's, and EFF pushes to +22.4\ensuremath{\pm}2.4\% at +8.9\%. Second, the
SLO column: both vendor profiles miss the tail SLO in a fraction of agentic repetitions (96\% compliance), while every one of our modes holds it in every repetition on both workloads. We attribute the
contrast to the acceptance gate of Algorithm 2, which admitted every setpoint only after it held ITL-p99 under serving-parity load, whereas a class-level profile is never tested against this
deployment's tail. The runtime guard (Algorithm 1, lines 7-9) did not trip in any of these runs, so the study shows what latency-gated calibration alone achieves, and H3's guard clause is established
by contract rather than by observation (\S{}6).

\subsection{The operating-mode ladder is monotone and SLO-guarded}\label{the-operating-mode-ladder-is-monotone-and-slo-guarded}

The three \emph{(ours)} rows of the \S{}5.1 table are the operating-mode ladder. Figure 6 plots the standard-workload ladder (235B, production calibrator revision; baseline 1.20\ensuremath{\pm}0.03
tokens/J at 896 W).

\begin{figure}[tbp]
\centering
\pandocbounded{\includegraphics[keepaspectratio,alt={The operating-mode ladder on the standard workload (235B). Efficiency gain (a) and mean node power saved (b) both increase monotonically down the ladder, every mode passes the ITL-p99 SLO, and the vendor profile's operating point (dashed) falls strictly inside the ladder.}]{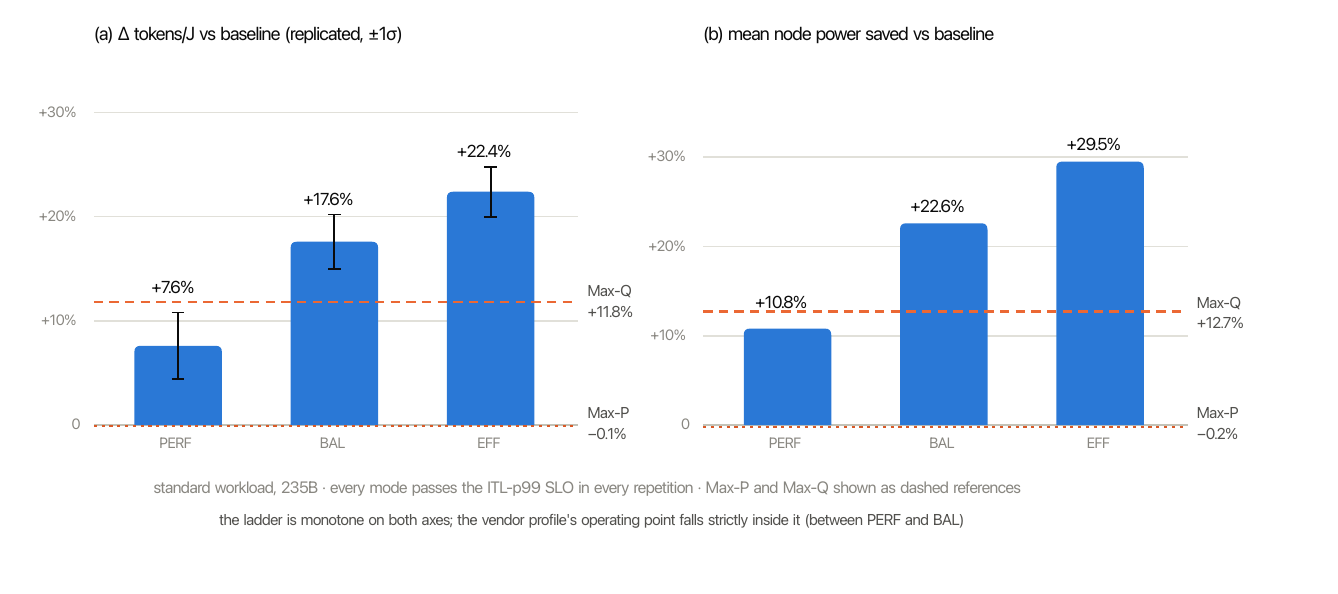}}
\caption{The operating-mode ladder on the standard workload (235B). Efficiency gain (a) and mean node power saved (b) both increase monotonically down the ladder, every mode passes the ITL-p99 SLO,
and the vendor profile's operating point (dashed) falls strictly inside the ladder.}
\end{figure}

On the standard workload the ladder is monotone on both axes, efficiency +7.6\% \ensuremath{\rightarrow} +17.6\% \ensuremath{\rightarrow} +22.4\% and mean power saved +10.8\% \ensuremath{\rightarrow}
+22.6\% \ensuremath{\rightarrow} +29.5\% from PERF to EFF, and Max-Q's operating point lies strictly inside it, between PERF and BAL on efficiency. On the agentic workload PERF is indistinguishable
from baseline, so monotonicity there is established for the BAL \ensuremath{\rightarrow} EFF step (+15.8\% \ensuremath{\rightarrow} +38.8\%). The mean-e2e increases down the ladder are the price of
the additional efficiency, and the correct lens for them is the latency \emph{budget}: every rung keeps ITL-p99 within the SLO, so the ladder spends headroom the SLO already grants, and the operator
chooses how much of it to convert. The agentic ladder rises far steeper because the agentic mix leaves more latency headroom to spend; the appropriate mode is workload-dependent (agentic
\ensuremath{\rightarrow} EFF, standard \ensuremath{\rightarrow} BAL), and the existence of this dependence argues for a ladder over a single profile. Figure 7 shows the controller's runtime behavior
at the setpoint level under a load transient.

\begin{figure}[tbp]
\centering
\pandocbounded{\includegraphics[keepaspectratio,alt={Runtime behavior under EFF (480B, agentic; 100 ms telemetry), with a 60 s load spike of +64 concurrency injected mid-run (shaded). Top: the prefill lane holds its in-window operating clock (1155 MHz) through the spike. Middle: decode-lane draw rides the commanded 500 W cap --- the cap binds continuously, spike included. Bottom: rolling ITL p99 (5 s window) sits at 24--28 ms in steady state, is pushed across the 30 ms SLO line during the spike (peak 96 ms), and returns to baseline within seconds of the spike ending; the run's aggregate ITL-p99 is 29.5 ms, inside the SLO. This capture isolates the steady-state contract --- window held, cap binding --- through a load transient, so the harness left the runtime guard of 3.2 out of the loop and the trace shows the crossing-and-recovery the guard is designed to bound. Guard behavior under load is the subject of 5.6, where the SLO sets the usable ceiling. Complements the run-level lane timeline of Figure A4.}]{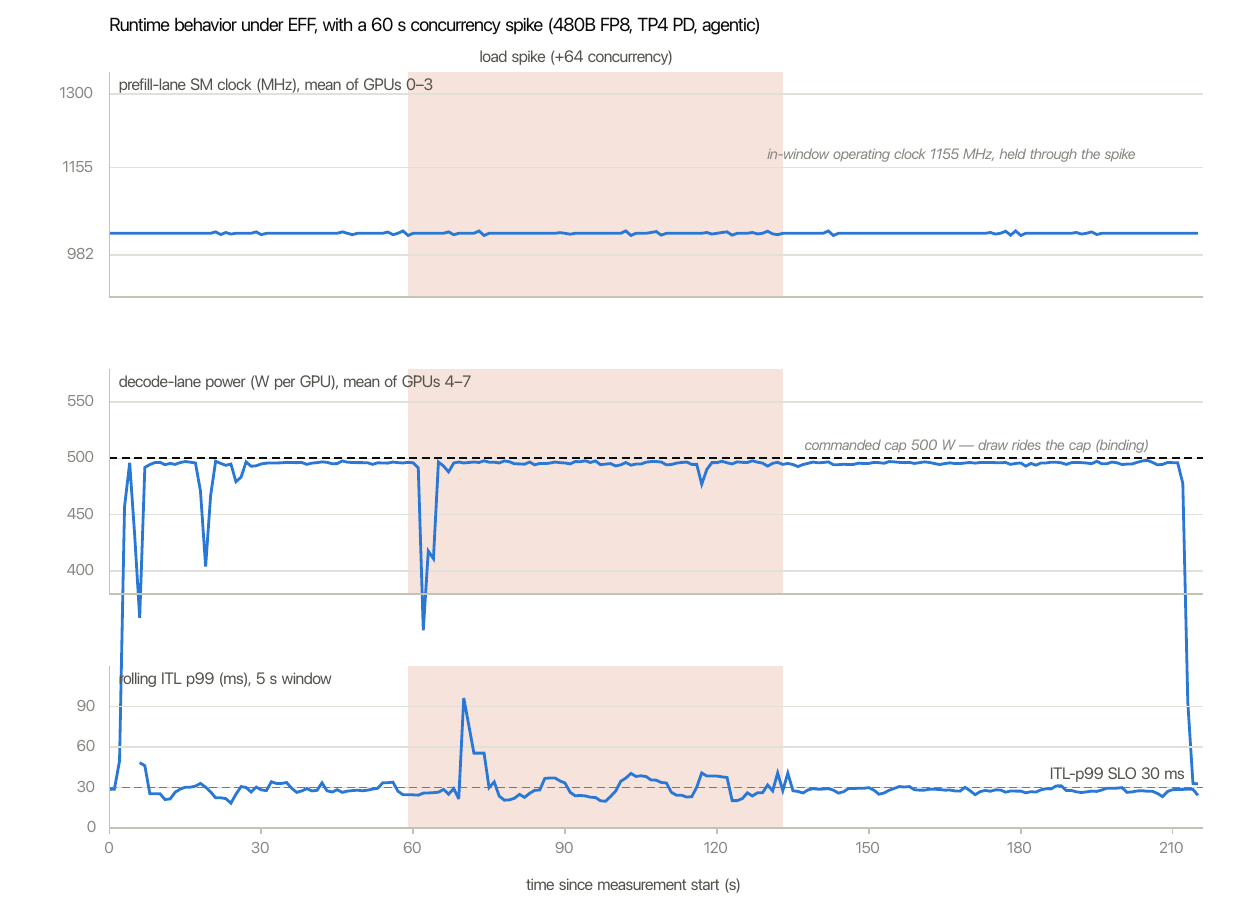}}
\caption{Runtime behavior under EFF (480B, agentic; 100 ms telemetry), with a 60 s load spike of +64 concurrency injected mid-run (shaded). Top: the prefill lane holds its in-window operating clock
(1155 MHz) through the spike. Middle: decode-lane draw rides the commanded 500 W cap --- the cap binds continuously, spike included. Bottom: rolling ITL p99 (5 s window) sits at 24--28 ms in steady
state, is pushed across the 30 ms SLO line during the spike (peak 96 ms), and returns to baseline within seconds of the spike ending; the run's aggregate ITL-p99 is 29.5 ms, inside the SLO. This
capture isolates the steady-state contract --- window held, cap binding --- through a load transient, so the harness left the runtime guard of \S{}3.2 out of the loop and the trace shows the
crossing-and-recovery the guard is designed to bound. Guard behavior under load is the subject of \S{}5.6, where the SLO sets the usable ceiling. Complements the run-level lane timeline of Figure A4.}
\end{figure}

\subsection{Decode actuator A/B: power cap beats static clock locking}\label{decode-actuator-ab-power-cap-beats-static-clock-locking}

{\begin{longtable}[]{@{}llllll@{}}
\toprule\noalign{}
Workload & Decode arm & \ensuremath{\Delta}tok/J & \ensuremath{\Delta}tok/s & \ensuremath{\Delta}decode power & ITL-p99 SLO \\
\midrule\noalign{}
\endhead
\bottomrule\noalign{}
\endlastfoot
agentic & \textbf{calibrated power cap (ours)} & \textbf{+24.1\%} & -8\% & -16\% & pass \\
& clock lock, throughput-optimal & +17.2\% & -9\% & -9\% & pass \\
& clock lock, aggressive & +15.5\% & -31\% & -37\% & pass \\
standard & \textbf{calibrated power cap (ours)} & \textbf{+18.8\%} & -12\% & -20\% & pass \\
& clock lock, throughput-optimal & +15.4\% & -3\% & -5\% & pass \\
& clock lock, aggressive & +8.8\% & -33\% & -37\% & pass \\
\end{longtable}
}

\begin{figure}[tbp]
\centering
\pandocbounded{\includegraphics[keepaspectratio,alt={Decode-lane actuator A/B on 480B (all arms share the same calibrated prefill efficiency floor; only the decode lever differs). Power cap wins on efficiency in both workloads while retaining throughput.}]{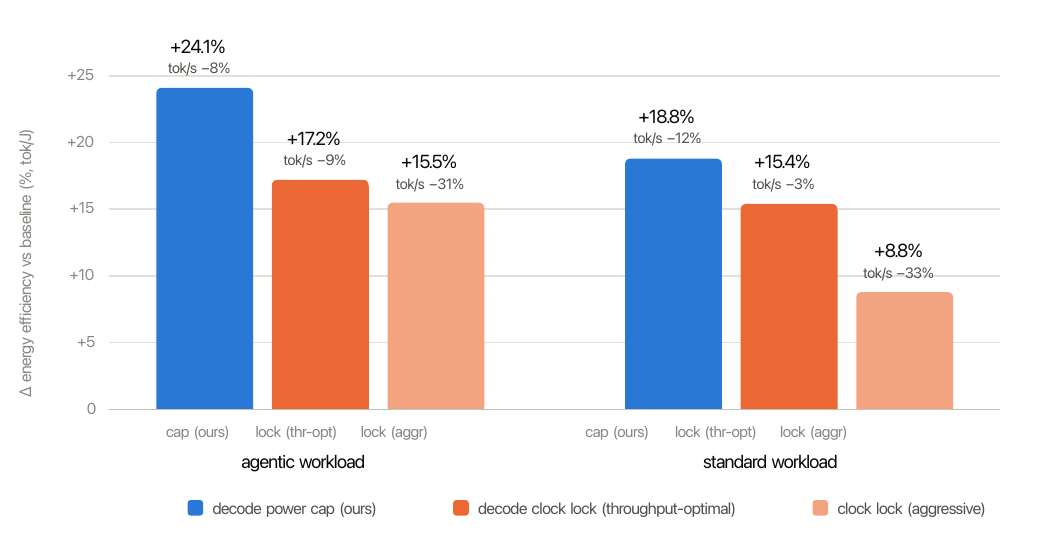}}
\caption{Decode-lane actuator A/B on 480B (all arms share the same calibrated prefill efficiency floor; only the decode lever differs). Power cap wins on efficiency in both workloads while retaining
throughput.}
\end{figure}

The cap arm is the calibrated EFF cap of \S{}3.3, placed below the lane's natural draw at the latency-gated cliff. The two lock arms bracket the static-clock design space: the
\emph{throughput-optimal} lock is the highest-throughput static clock, the strongest case for POLCA-style locking, and the \emph{aggressive} lock trades throughput for the deepest power reduction; the
exact lock frequencies are sweep outputs for this stack and carry no meaning beyond these two roles. All arms share the calibrated prefill floor, so the \ensuremath{\Delta}tok/J column includes the
prefill contribution and the comparison between rows isolates the decode lever. The ranking cap \textgreater{} lock-optimal \textgreater{} lock-aggressive holds in both workloads. The aggressive lock
saves power but collapses throughput (-31\% to -33\%); the throughput-optimal lock preserves throughput but barely cuts decode power (-5\% to -9\%), so its efficiency gain is mostly inherited from the
shared prefill floor. The cap achieves both because the GPU self-DVFSes under it: peak power is bounded while clock allocation remains adaptive. A static frequency cannot occupy both ends of a
flat-then-cliff curve; a cap can, by construction.

\textbf{What this does and does not refute.} POLCA's Insight 7 rejected capping as reactive and unpredictable, based on colocated A100 serving with prompt spikes, measured through a throughput lens.
On a disaggregated decode lane there are no spikes to overshoot, the calibrated cliff makes the cap's performance effect deterministic, and POLCA's chosen mechanism, the static clock lock, is the arm
that loses the A/B; POLCA's own caveat, that locks tax execution constantly, is the cost we avoid by confining clock control to the prefill lane. The A/B therefore refutes the frequency-first position
\emph{for static locking on disaggregated decode lanes}. It does not test adaptive per-iteration DVFS of the kind VoltanaLLM and DualScale implement \cite{voltana,dualscale}; that comparison would
need their controllers running on our stack under our SLO gate (\S{}6). The characterization in \cite{illusion}, which reports clock locking dominating capping in decode, is reconciled by cap
placement: a cap at or near TDP on a memory-bound lane never binds and yields no savings. Our caps bind by construction (\ensuremath{\approx}85\% of natural draw on 480B and 80\% on 235B, at a cliff
jointly gated on throughput and ITL-p99), and the studies differ in kind, theirs an open-loop characterization across attention architectures, ours a closed-loop serving A/B with total-node energy
accounting and SLO gates. The two are consistent measurements of different operating points, uncalibrated caps and calibrated ones, which is itself evidence that \emph{calibration, rather than
mechanism choice alone, is the load-bearing component}.

\textbf{Binding fraction.} The load-axis campaign (\S{}5.6) measured how often the cap actually binds. Across the concurrency range the decode cap is engaged 94-99\% of the time under the standard
workload, and only 13-24\% of the time under the agentic workload above concurrency 32, with the same controller and setpoint; the efficiency advantage tracks that difference. We report the
association and stop there: the binding fraction is an outcome of the treatment and the realized load, not an independent variable. A probe underlines the caution: removing the think gaps from the
agentic workload was expected to raise binding by removing idle time, but binding fell from 55\% to 23.5\% and the gain from +18.1\% to +6.1\%, because gap-free workers synchronize onto turn
boundaries and the lane alternates between prefill and decode waves that keep decode draw below the cap. What sets binding is the phase-overlap structure, not the idle fraction.

\subsection{Calibration robustness: the repeated-calibration study}\label{calibration-robustness-the-repeated-calibration-study}

Repeated calibrate-and-measure cycles on 480B exposed two failure classes that any deployable calibrator must fix, and that to our knowledge no prior power-management evaluation reports.

\begin{enumerate}
\def\labelenumi{\arabic{enumi}.}
\tightlist
\item
  \textbf{Setpoint instability on flat curves.} Decode's throughput-vs-cap curve is flat above the cliff, so small noise in the reference against which candidates are judged translates into large
  shifts in the estimated cliff. An early calibrator revision produced decode setpoints from \ensuremath{\approx}79\% of natural draw up to effectively uncapped across runs (bimodal, most near
  \ensuremath{\approx}82\% of natural with occasional collapses to no-cap; Figure 9). The production calibrator resolves this with the multi-repetition, noise-robust reference statistic that Algorithm
  2 (line 1) requires, and its output is stable across the final ladder (\S{}5.2). Estimating a crossing point on a flat curve is inherently noise-fragile, and reporting single-run cliff estimates
  without a reproducibility study is unsafe.
\item
  \textbf{Throughput-blind acceptance violates SLOs.} The 235B EFF cap accepted on throughput alone (\S{}3.3) passed throughput while violating the ITL-p99 SLO in the large majority of runs; the
  latency-aware criterion restored full compliance at +22.4\% tokens/J. \emph{Any} setpoint chosen by throughput alone, a calibrated cap, a locked clock, or a vendor profile, can hide a tail-latency
  violation.
\end{enumerate}

\begin{figure}[tbp]
\centering
\pandocbounded{\includegraphics[keepaspectratio,alt={Accepted EFF decode cap across 30 back-to-back independent calibrations of the same (model, quant, engine) combination, early calibrator revision (480B). The estimates cluster at 460--500 W (21/30) but scatter to 540--808 W and collapse to no-cap twice --- the signature of estimating a crossing point on a flat curve from single-run reference measurements. The production revision's noise-robust acceptance settles at 520 W on the same stack (dashed). Single-run cliff estimates reported without a reproducibility study are unsafe (failure class 1).}]{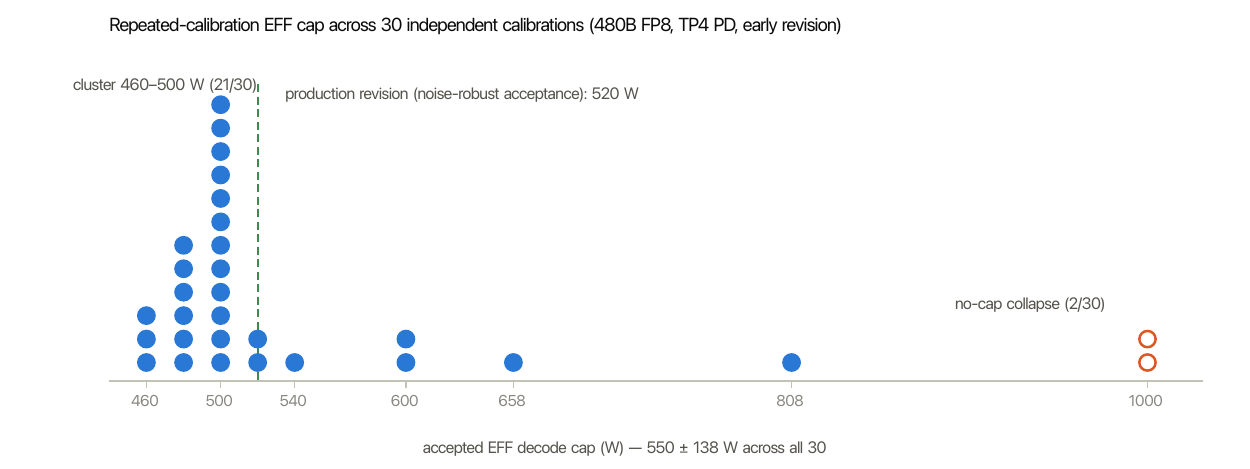}}
\caption{Accepted EFF decode cap across 30 back-to-back independent calibrations of the same (model, quant, engine) combination, early calibrator revision (480B). The estimates cluster at 460--500 W
(21/30) but scatter to 540--808 W and collapse to no-cap twice --- the signature of estimating a crossing point on a flat curve from single-run reference measurements. The production revision's
noise-robust acceptance settles at 520 W on the same stack (dashed). Single-run cliff estimates reported without a reproducibility study are unsafe (failure class 1).}
\end{figure}

Robustness of the primary results is otherwise good: paired-\ensuremath{\Delta} statistics across the repeated runs are tight (standard Max-Q +11.8\ensuremath{\pm}2.0\%, our BAL
+17.6\ensuremath{\pm}2.6\%), and the intermittent TTFT blowups are arm-independent stack behavior, dropped pairwise, with root cause tracked separately.

\textbf{Where the power savings come from: per-lane attribution.} The repeated study's lane-level telemetry decomposes the node power saving by lane. At EFF on 480B, the prefill lane contributes the
bulk of the reduction, lane power -43\% (standard) and -47\% (agentic), while the calibrated decode cap contributes -16\% and -14\% respectively, for node totals of -26\% and -28\%. The asymmetry is
consistent with Equation (1): prefill power tracks \(f^{1+\alpha}\), so a window floor purchases a large power reduction, whereas decode already draws flat and well below TDP, leaving the cap a
thinner but latency-cheap margin. The two mechanisms also differ in output stability: the prefill window floor was identical in every one of the repeated calibrations, while the decode cap carried all
of the run-to-run variance (failure class 1). Two attributions this telemetry cannot make must be stated. It attributes \emph{power}, not \emph{efficiency}: a factorial ablation (prefill-window-only /
decode-cap-only / both, each against BL on both workloads, reporting the \S{}4 metric triple) was not carried and remains outstanding. And because the in-window runtime controller was active in every
arm (\S{}3.1), the prefill-lane figure bundles the window floor with the in-window policy; a floor-only arm (window pinned at its floor, controller inactive) would separate the two.

\subsection{Composability: substitutes, not complements}\label{composability-substitutes-not-complements}

A 4-arm study (BL / Max-Q / ours / ours+Max-Q) on 480B indicated that once our controller governs the lanes, additionally enabling Max-Q yields no further efficiency: the vendor profile and our
controller act on the same power headroom, and ours captures it at a more favorable efficiency-latency exchange. The mechanisms are therefore alternatives rather than complements, and the Pareto
comparison of \S{}5.1 is the decision criterion.

\subsection{Load dependence: the advantage across the concurrency range}\label{load-dependence-the-advantage-across-the-concurrency-range}

The results above are measured at one concurrency per workload. To characterize the load axis we swept nine concurrency points from 4 to 96 on 235B, with the engine's maximum running requests raised
from 16 to 64 so that client concurrency, not the engine queue, sets the offered load. Arms are BL, the vendor Max-Q profile and EFF, paired block-by-block within a session, three repetitions per
cell.

\begin{figure}[tbp]
\centering
\pandocbounded{\includegraphics[keepaspectratio,alt={Efficiency advantage of EFF over the shipped Max-Q profile across nine concurrency points (235B NVFP4, TP1 PD; block-paired  tokens/J, bars 1). On the standard workload the advantage holds at every point and widens with load; on the agentic workload it is real but flat and thin. Above the shaded boundary EFF no longer holds the ITL-p99 SLO, so the widening advantage there is not adoptable under an SLO-bound frame.}]{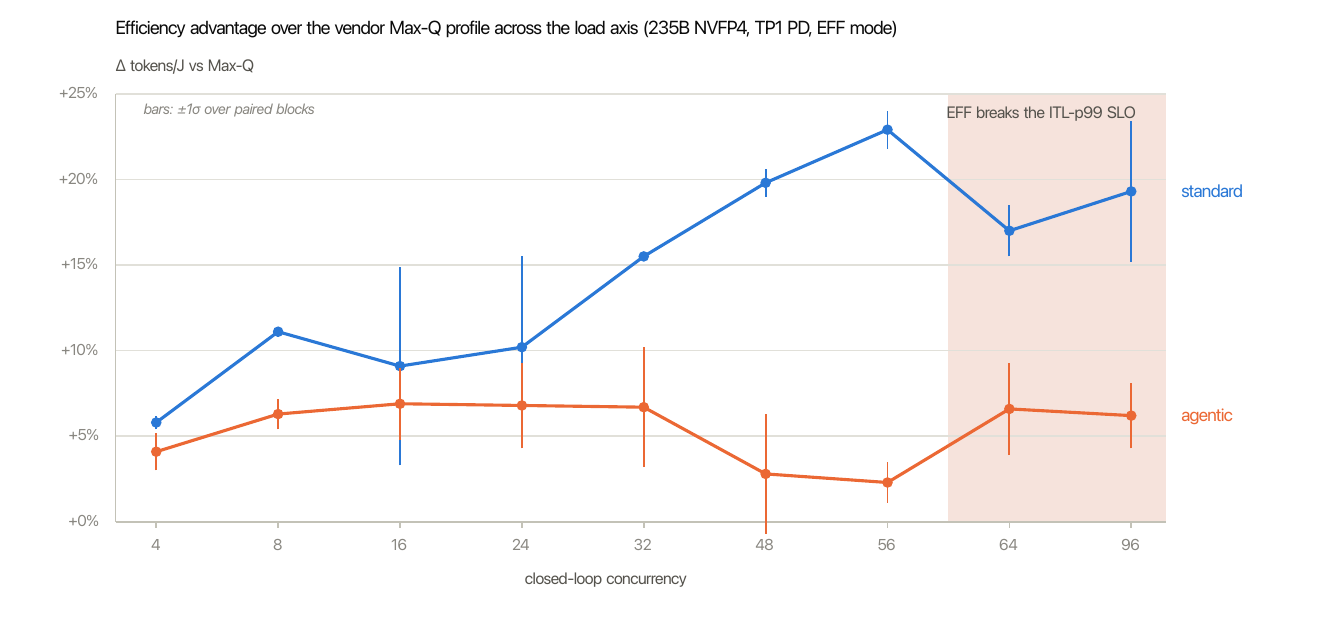}}
\caption{Efficiency advantage of EFF over the shipped Max-Q profile across nine concurrency points (235B NVFP4, TP1 PD; block-paired \ensuremath{\Delta} tokens/J, bars
\ensuremath{\pm}1\ensuremath{\sigma}). On the standard workload the advantage holds at every point and widens with load; on the agentic workload it is real but flat and thin. Above the shaded boundary
EFF no longer holds the ITL-p99 SLO, so the widening advantage there is not adoptable under an SLO-bound frame.}
\end{figure}

\textbf{On standard traffic the advantage holds across the whole range and grows with load}, from +5.8\ensuremath{\pm}0.4\% at concurrency 4 to +22.9\ensuremath{\pm}1.1\% at 56; against the
uncontrolled baseline the same sweep rises from +15.8\% at concurrency 4 into roughly the +30\% band from 32 upward. We do not name a peak: two clean sessions at concurrency 64 disagreed by 6.0
percentage points, and two further sessions established +30.6\ensuremath{\pm}1.8\% as the representative value, which leaves neither ``the gain peaks at 64'' nor ``it declines by 96'' supportable. The
session-to-session disagreement is also a caution about the \ensuremath{\pm}1\ensuremath{\sigma} figures elsewhere in this paper, which are within-session dispersions.

\textbf{On agentic traffic the advantage is real but thin}, between +2.3\% and +6.9\% at every point, with no trend. The reason is in the load rather than the controller: completed requests on the
agentic baseline are flat from concurrency 24 to 96 (595 \ensuremath{\rightarrow} 612 \ensuremath{\rightarrow} 596 per 180 s) while mean power falls (735 \ensuremath{\rightarrow} 609 W), because the
think gaps cap the arrival rate. Past roughly concurrency 24-32 the x axis measures waiting workers rather than demand, and the decode lane spends most of its time below the cap (\S{}5.3).

\textbf{Mode ordering is stable across load.} In all twelve workload-concurrency cells the point estimates order as EFF \textgreater{} BAL \textgreater{} PERF \ensuremath{\geq} Max-Q. The cells share
sessions and baselines, so this is one consistent pattern rather than twelve confirmations, and we do not read differences within \ensuremath{\pm}1.3 percentage points, the largest ordering effect
between arms, as a distinguished advantage.

\textbf{The SLO sets the usable ceiling, and it is below the efficiency ceiling.} EFF holds ITL-p99 at concurrency 48 (two sessions, 6/6 runs, 27.4-28.7 ms) and 56 (one session, 3/3, 28.1-28.6 ms),
and breaks it at 64 (two sessions, 6/6, 30.2-32.8 ms) and 96, so the compliant ceiling falls in {[}56, 64) on this grid. Efficiency keeps climbing above it (concurrency 64 measured +36.7\% against
baseline in one session), but an operator bound by the tail SLO cannot spend that: the efficiency number alone would have recommended a setting the latency axis forbids. The bracket is conditional on
this grid and on a healthy stack; from concurrency 48 upward we observed intermittent router-circuit episodes affecting all arms, including the baseline.

\subsection{Sustained operation: what a day of running actually saves}\label{sustained-operation-what-a-day-of-running-actually-saves}

Every number above is a ratio over a run of a few minutes. The operational question is how much electricity a day of serving costs, which requires establishing that short-run ratios survive a day of
thermal and workload reality. We ran the three arms \textbf{in parallel on one node}, each on its own GPU pair (Max-Q, EFF, and the uncontrolled baseline), under the standard workload at concurrency
24, in 96 segments of 900 s per day, and repeated the whole day three times. Running the arms simultaneously removes time-of-day drift from the comparison by construction; the cost is that the three
arms share the host's CPU, PCIe fabric and cooling, and we did not isolate cross-arm coupling through those shared resources. Energy is integrated from 1 Hz telemetry \textbf{over each arm's two
GPUs}, so these absolute values are not comparable with the eight-GPU accounting used elsewhere.

{\begin{longtable}[]{@{}lllll@{}}
\toprule\noalign{}
Arm & Energy (kWh/day) & Saved vs baseline & Saved vs baseline (\%) & \ensuremath{\Delta} tokens/kJ \\
\midrule\noalign{}
\endhead
\bottomrule\noalign{}
\endlastfoot
Uncontrolled baseline & 22.87 \ensuremath{\pm} 0.04 & --- & --- & --- \\
Max-Q & 19.26 \ensuremath{\pm} 0.03 & 3.62 \ensuremath{\pm} 0.00 kWh & 15.8 \ensuremath{\pm} 0.2\% & +12.6 \ensuremath{\pm} 0.2\% \\
\textbf{EFF (ours)} & \textbf{15.49 \ensuremath{\pm} 0.05} & \textbf{7.39 \ensuremath{\pm} 0.02 kWh} & \textbf{32.3 \ensuremath{\pm} 0.2\%} & \textbf{+19.5 \ensuremath{\pm} 0.4\%} \\
\end{longtable}
}

Over three independent days the controller saved \textbf{7.39 \ensuremath{\pm} 0.02 kWh/day, 32.3\% of what the uncontrolled lane pair consumed}, and 3.77 \ensuremath{\pm} 0.03 kWh/day more than the
vendor profile. The run-to-run standard deviation of 0.02 kWh/day is the tightest reproducibility figure in this paper. Max-Q's +12.6\% tokens/kJ here is measured under this campaign's two-GPU
accounting and 900 s windows and is not the same quantity as its +11.8\% in the \S{}5.1 table.

\textbf{Nothing drifted.} Decode natural draw held at 688-689 W, the cap stayed engaged 95-97\% of the time, the prefill lane held its in-window operating point at 1230 MHz, and decode temperature
settled at 45-52 \textdegree{}C, all flat across 24 hours, three times over. Thermal equilibrium is reached within the first hour and the calibrated setpoints remain valid beyond it. The tail SLO held
in \textbf{all 864 segments}, and the runtime guard never had cause to fire.

\textbf{Short-run ratios extrapolate; short-run absolute energy does not.} Projecting one hour to a full day over-predicted absolute energy in every arm of every run, by 1.8-10.1\%, and the sign never
flipped: the first hour is a cold machine drawing more for the same work. The \emph{difference between arms} extrapolated within -6.6 to +3.0\% without a systematic direction. An operator can carry a
short-run \ensuremath{\Delta}\% into a capacity estimate, but must measure to state kilowatt-hours.

Two operational notes. The shipped Max-Q profile is not a static cap: the enforced limit sits at 1000 W and moves intermittently within 567-994 W, which is part of why a static clock lock is not a
substitute for it (\S{}4). And at the end of the first day the command clearing the vendor profile silently did nothing (it requires the profile list as an argument) and the profile stayed resident;
the between-run check caught it, the clear command was pinned and followed by a read-back, and the two subsequent days verified clean. Leaked power state is exactly the failure that silently
invalidates the run that follows it.

\section{Discussion}\label{discussion}

\textbf{Scope vs POLCA.} POLCA solves cluster-level power oversubscription (fast reclaim when a breaker threshold is breached); we solve steady-state lane efficiency, and the rebuttal in \S{}5.3 is
scoped accordingly. For the reclaim use case, a calibrated cap is a deterministic, pre-validated reclaim point, and B200-generation in-band capping latency is far below the 40 s out-of-band budget
POLCA designed around; quantifying reclaim latency under our controller is future work.

\textbf{Hardware and model generality.} All results are single-node B200 on two Qwen3-family MoE models (480B FP8, 235B NVFP4) under one engine stack. The \emph{mechanism} argument (flat memory-bound
decode power; self-DVFS under caps) is not Blackwell-specific, but cliff positions move across GPU generations, which is an argument \emph{for} automatic calibration rather than against the design.
The mechanism is architecture-specific by construction: MoE's per-token expert sparsity leaves the decode lane below its power ceiling, which is exactly the slack a calibrated cap reclaims. Appendix A
found this margin roughly 5\ensuremath{\times} smaller on a comparably sized dense model (+5.7\ensuremath{\pm}8.0\% vs +26.0\ensuremath{\pm}6.4\% cap-only tokens/J), because dense models saturate the
tensor pipes and draw near the device ceiling. We make no dense-model claims; a controlled dense-model evaluation under the full controller is left to future work.

\textbf{Datacenter implications.} In a power-constrained facility, the tokens a fixed provisioned budget can deliver scale with energy efficiency, so the gains of \S{}5 translate into capacity:
+17.6-22.4\% more SLO-compliant tokens per provisioned megawatt on the standard workload and +15.8-38.8\% on the agentic workload, against +11.4-11.8\% for the vendor profile on the same model. Mean
node power reductions compound this (a -22.6\% mean draw admits roughly 29\% more serving instances under the same budget), and \S{}5.7 puts an absolute figure on one lane pair: 7.39 \ensuremath{\pm}
0.02 kWh saved per day against no control, 3.77 kWh against the vendor profile. The cap-based decode lane adds a property scheduling-based savings lack: the per-GPU peak is bounded
\emph{deterministically} by the commanded cap, the quantity provisioning and oversubscription planning actually consume.

\begin{figure}[tbp]
\centering
\pandocbounded{\includegraphics[keepaspectratio,alt={Facility-level translation of the efficiency results: SLO-compliant tokens deliverable per provisioned megawatt, indexed to baseline = 100 (235B, paired repeated runs). The vendor profile buys roughly 11 points of capacity; per-stack calibrated modes buy 22 to 39 depending on workload.}]{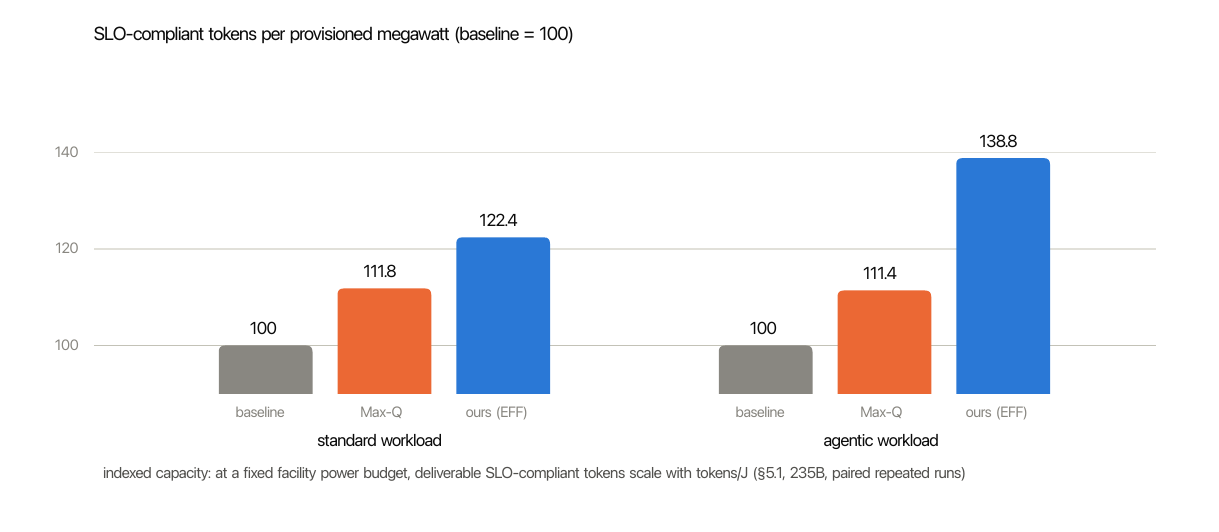}}
\caption{Facility-level translation of the efficiency results: SLO-compliant tokens deliverable per provisioned megawatt, indexed to baseline = 100 (235B, paired repeated runs). The vendor profile
buys roughly 11 points of capacity; per-stack calibrated modes buy 22 to 39 depending on workload.}
\end{figure}

\textbf{Limitations.} (i) A residual efficiency-latency trade remains on the standard workload (up to +8.9\% mean-e2e at EFF); the ladder makes it explicit and selectable rather than hidden. (ii) The
intermittent TTFT anomalies in our baseline (\textasciitilde15\% of repetitions) are a serving-stack artifact whose root cause remains open. (iii) The scope is a single node: cross-node power shifting
and cluster-level scheduling (as in DynamoLLM \cite{dynamollm}) are orthogonal layers that compose above per-lane control. (iv) Calibration time is the principal operational cost: a vendor profile
applies in minutes, whereas our pass takes about an hour per (model, quantization, engine, PD configuration) and re-runs whenever the fingerprint changes; at serving scale it amortizes within hours,
but class-level profiles do not carry it. (v) The load and time axes are characterized only on 235B; the 480B results are single-point, and the SLO-compliant ceiling is a bracket on one grid under a
healthy stack. (vi) The decode A/B compares the cap against static clock locks only; adaptive per-iteration DVFS controllers \cite{voltana,dualscale} were not run on our stack, so the claim that
capping beats frequency control is established against static locking, not against those systems. (vii) The prefill-lane saving bundles the window floor with a proprietary in-window controller, and
the factorial per-lane ablation that would attribute the efficiency gain by mechanism remains outstanding (\S{}5.4). (viii) Both workloads use closed-loop arrivals; external validity under native
trace arrival processes is untested.

\section{Related Work}\label{related-work}

\textbf{Disaggregated serving substrate.} Splitwise \cite{splitwise} established phase splitting (1.4\ensuremath{\times} throughput at 20\% lower cost) and characterized the phases' distinct power
profiles, observing that decode does not need the latest GPUs' full capability: power differentiation at provisioning time, which our runtime-calibrated, SLO-gated actuation generalizes. DistServe
\cite{distserve} and Mooncake \cite{mooncake} optimize goodput without power actuation. These systems are the substrate on which our work operates rather than points of comparison.

\textbf{Vendor static profiles.} NVIDIA's workload power profiles \cite{maxq} are the deployable baseline, and their own data shows multi-knob recipes including a power limit beating frequency-only
scaling by 7-9\%. We compare against the real in-flight profiles and show a calibrated, phase-decoupled ladder Pareto-dominates them on the agentic workload (\S{}5.1).

\textbf{Cloud LLM power management.} POLCA \cite{polca} supplied the phase power characterization we build on and concluded that frequency capping is the reliable control; \S{}5.3 shows this
conclusion inverts for static locking on disaggregated decode lanes. TAPAS \cite{tapas} manages thermal and power via placement, routing and reconfiguration, and RAPID \cite{rapid} does power-aware
prefill/decode role assignment; both are cluster-level scheduling levers orthogonal to and composable with per-lane actuation.

\textbf{Phase-aware DVFS controllers.} VoltanaLLM \cite{voltana} is the closest system work: feedback-driven per-phase frequency control plus state-space routing in P/D-disaggregated SGLang, up to
36.3\% energy savings. DualScale \cite{dualscale} co-designs phase-aware placement with per-iteration DVFS (ILP re-provisioning every five minutes, then MPC for prefill and slack-aware stepping for
decode; up to -39\% prefill and -48\% decode energy vs DistServe on 16\ensuremath{\times} H100). Its own decomposition is instructive: the coarse tier, which chooses instance counts, parallelism,
routing weights \emph{and} a static baseline frequency per instance, captures most of the saving, while per-iteration DVFS adds between -4\% and +20\% on top. A well-chosen static setpoint is where
the energy is, which is the premise of our calibration and mode ladder, and that static frequency is exactly the quantity we replace with a calibrated cap on the decode lane. DualScale also trains its
models per (model, GPU) pair without an engine-stack fingerprint (\S{}2.3). GreenLLM \cite{greenllm} fits latency-power models over SM frequency per phase for colocated serving; throttLL'eM
\cite{throttllem} predictively throttles frequency phase-uniformly; DynamoLLM \cite{dynamollm} reconfigures instance count, parallelism and frequency per load epoch from offline profiles. All five
actuate through frequency; none compares frequency against calibrated power caps on the decode lane, calibrates per engine stack, or evaluates against the in-flight vendor profile. Our work composes
with them at the scheduling layer and differs on the actuation mechanism; a direct comparison against their adaptive DVFS is the open experiment of \S{}6.

\textbf{Power capping for inference.} PALS \cite{pals} treats NVML power caps plus batch size as joint knobs for MoE serving in vLLM (up to 26.3\% efficiency, 4-7\ensuremath{\times} fewer QoS
violations); ICPP'25 work \cite{icpp-cap} argues caps should be a first-class inference-runtime primitive. Both operate on monolithic serving without phase decoupling. Conversely, a 2026
characterization \cite{illusion} argues decode power capping is an illusion that clock locking dominates; \S{}5.3 reconciles this with our opposite finding through cap placement.

\textbf{Training energy.} Zeus \cite{zeus} tunes power limits and batch size for training; Perseus \cite{perseus} removes energy bloat along pipeline critical paths, the training-side analogue of
phase decoupling; EnvPipe \cite{envpipe} exploits pipeline bubbles. These motivate our exclusion of training (time-synchronized phases, no spatial lanes) rather than competing with it.

\textbf{Positioning.} No prior system (i) selects the actuation \emph{mechanism} per disaggregated lane from measurement, (ii) derives cap and clock setpoints from an automatic per-(model,
quantization, engine-stack) calibration with a latency-gated cliff and an engine fingerprint, or (iii) evaluates power control directly against the shipped vendor profile on the mean-e2e + tail-ITL
plane. Phase-awareness alone ceased to be novel in 2025; per-lane mechanism selection, stack-tracking calibration, and latency-inclusive evaluation are the contributions of this paper.

\section{Conclusion}\label{conclusion}

This work began from the gap between the reported and the realized efficiency of a shipped vendor power profile on disaggregated LLM serving. Evaluated in-flight against NVIDIA Max-Q on B200 hardware,
a phase-decoupled, per-stack-calibrated operating-mode ladder achieves 1.5-3.4\ensuremath{\times} the vendor profile's efficiency gain, Pareto-dominating it on agentic workloads, and every
latency-gated operating point holds the tail SLO in every repetition where the static vendor profiles miss it in a fraction of runs. The measurements also extended our hypotheses in one respect we had
not anticipated: the two lanes require different \emph{mechanisms}, not merely different settings. On decode lanes a calibrated power cap outperforms the static frequency locking recommended by prior
work, because disaggregation converts the phase-management problem from a temporal one into a spatial one and thereby restores the viability of an actuator the literature had set aside. The broader
lesson is methodological: power-control conclusions derived from intuition or validated solely through throughput (the vendor's, prior frameworks', and initially our own) were not sustained once an
end-to-end latency axis was applied.

\section*{Future Work}

Four experiments would close the gaps \S{}6 identifies: running an adaptive per-phase DVFS controller \cite{voltana} on our stack under our SLO gate as a fourth arm of the decode A/B; a
fault-injection study that forces the runtime guard to trip and measures its recovery time and cost; the factorial per-lane ablation with a floor-only prefill arm; and evaluation under native trace
arrival processes. Beyond these, a training-side comparison is deferred (training lacks the spatial phase separation our lane actuators exploit, phase transitions outrun VRM slew of
\textasciitilde12.5 ms, and Zeus/Perseus/EnvPipe already serve that regime), cross-node reclaim with calibrated caps (POLCA's oversubscription scenario) is planned, and we are working on the
approach's principal cost, the calibration pass: shortening its wall time and developing online calibration that refines setpoints from live serving telemetry under the same SLO guard.

\section*{Reproducibility}

Everything needed to reproduce every measured configuration is disclosed: the setpoints of every arm (\S{}5, relative to the lane's natural draw and maximum clock, the two anchors any deployment can
measure locally), the calibration acceptance contract (Algorithm 2), the runtime-guard contract (Algorithm 1), the source data behind the measured figures, and the measurement protocol (pairwise
drops, actuation-grid quantization, serving-parity load). Raw per-repetition JSON and the calibration artifacts used in each experiment, including their engine-stack fingerprints, are retained and
available on request.

Two components are described only at the level of their behavioral contract: the setpoint-search heuristic and the in-window clock policy, both part of our production system's control logic. Any
search whose outputs pass the acceptance gate of Algorithm 2 produces valid setpoints, and \S{}5.4 characterizes the output stability of ours. The in-window policy is a different case: the guarantees
of \S{}5 derive from the window bounds, but because the policy was active in every measured arm, its contribution to the prefill-lane saving is not separated from the floor's (\S{}5.4, \S{}6), and a
floor-only reproduction is the experiment that would separate them.

\appendix
\setcounter{figure}{0}
\renewcommand{\thefigure}{A\arabic{figure}}
\renewcommand{\theHfigure}{A\arabic{figure}}

\section{The actuator-selection campaign}\label{the-actuator-selection-campaign}

The per-lane design of \S{}3.1 (decode: power cap; prefill: clock window) and the choice of PD-disaggregated serving as the target emerged from an exploratory campaign, conducted before the main
evaluation, across four base models (a 70B dense, a 405B dense, a 480B MoE with 35B active parameters, and a 120B hybrid Mamba-Transformer MoE with 12B active), three quantizations (BF16, FP8, NVFP4),
three workload shapes (prefill-heavy, decode-heavy, mixed), and both colocated and PD-disaggregated topologies. This appendix summarizes the evidence for three questions: why target disaggregated
serving (A.2), why the power cap and why MoE models benefit most (A.3), and what phase-decoupled actuation does at run granularity (A.4). Consistent with \S{}3.3, all setpoints are reported relative
to the lane's natural draw (\(P_{\text{nat}}\)) or the maximum SM clock.

\subsection{Phase signatures are model-specific but topology-invariant}\label{phase-signatures-are-model-specific-but-topology-invariant}

Using per-GPU activity counters (tensor-pipe utilization vs DRAM utilization, sampled at 100 ms), prefill and decode form disjoint clusters for every model tested, and a threshold classifier derived
from these counters labels held-out samples with 100\% accuracy. The separation gap, however, varies by two orders of magnitude (Figure A1), and the same 70B model measured colocated and
PD-disaggregated produced statistically indistinguishable signatures (prefill counter-ratio 4.51 vs 4.48; decode 0.032 in both).

\begin{figure}[tbp]
\centering
\pandocbounded{\includegraphics[keepaspectratio,alt={Prefill/decode activity signatures per model (tensor-pipe : DRAM counter ratio, log scale). Every model separates cleanly, but the separation gap spans 2.7 to 141, so classification thresholds cannot be hard-coded; the signature is invariant to serving topology.}]{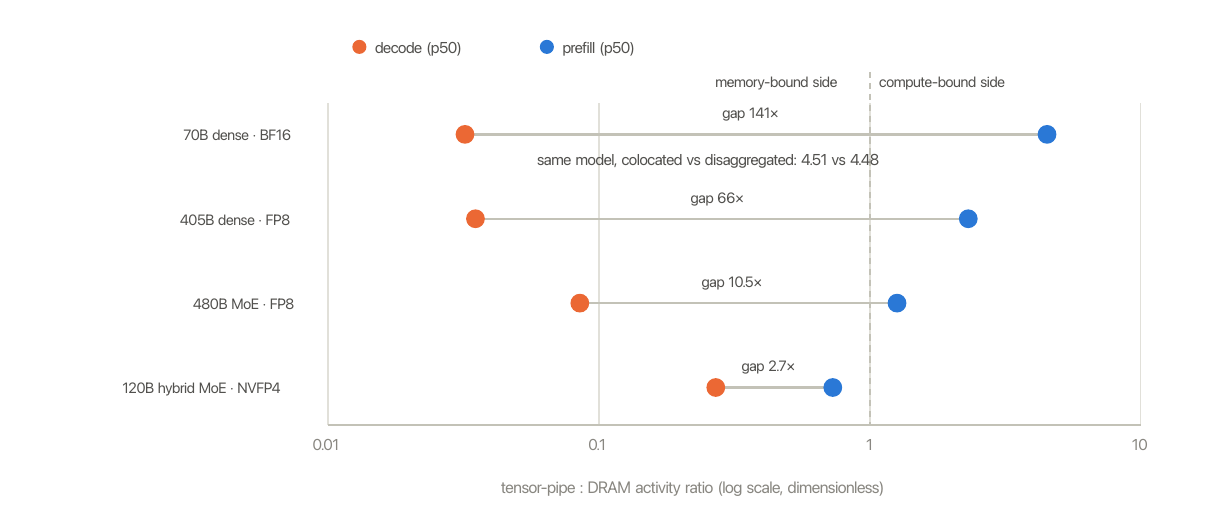}}
\caption{Prefill/decode activity signatures per model (tensor-pipe : DRAM counter ratio, log scale). Every model separates cleanly, but the separation gap spans 2.7\ensuremath{\times} to
141\ensuremath{\times}, so classification thresholds cannot be hard-coded; the signature is invariant to serving topology.}
\end{figure}

Two consequences carried into the main design: phase behavior is a property of the (model, quantization) pair rather than of the deployment topology, so a per-model characterization transfers across
topologies; and under disaggregation a lane's phase is static, which eliminates runtime phase detection entirely (\S{}2.1). The same counters also quantify how far each architecture sits from compute
saturation, which A.3 connects to the size of the reclaimable gain.

\subsection{Why disaggregated serving is the target}\label{why-disaggregated-serving-is-the-target}

In colocated serving one GPU carries both execution regimes, so it holds exactly one power setting, and that single setting faces a zero-sum trade between the two panels of Figure A2. A setting sized
for decode-phase efficiency wins panel (b) and loses panel (a): it is near-lossless within the decode phase, but the moment a prefill burst arrives on the same GPU, tail latency collapses (TTFT p99
+1005\% on a colocated MoE deployment). A setting relaxed enough to protect prefill wins panel (a) and loses panel (b): latency stays bounded, but roughly a third of the achievable decode-phase gain
is forgone (+18-20\% versus +28-31\% on the same deployment). Disaggregation removes the trade itself: each lane holds exactly one regime, statically (A.1), so each lane holds its own mechanism and
setpoint, and the rightmost column of Figure A2 is the only one that is best in both panels simultaneously. The main-body results confirm this; on the agentic workload, +38.8\% tokens/J arrives
together with a 5.8\% TTFT p95 improvement and full ITL-p99 SLO compliance (\S{}5.1).

\begin{figure}[tbp]
\centering
\pandocbounded{\includegraphics[keepaspectratio,alt={The colocated dilemma, and its structural resolution by disaggregation. Panels share the x axis: (a) prefill tail-latency cost and (b) energy-efficiency gain for the same three configurations. A single colocated setting either collapses prefill latency (left, axis broken at +1005\%) or forgoes a large fraction of the decode-phase gain (middle); per-lane mechanisms on disaggregated serving (right) obtain the largest efficiency gain while improving tail latency.}]{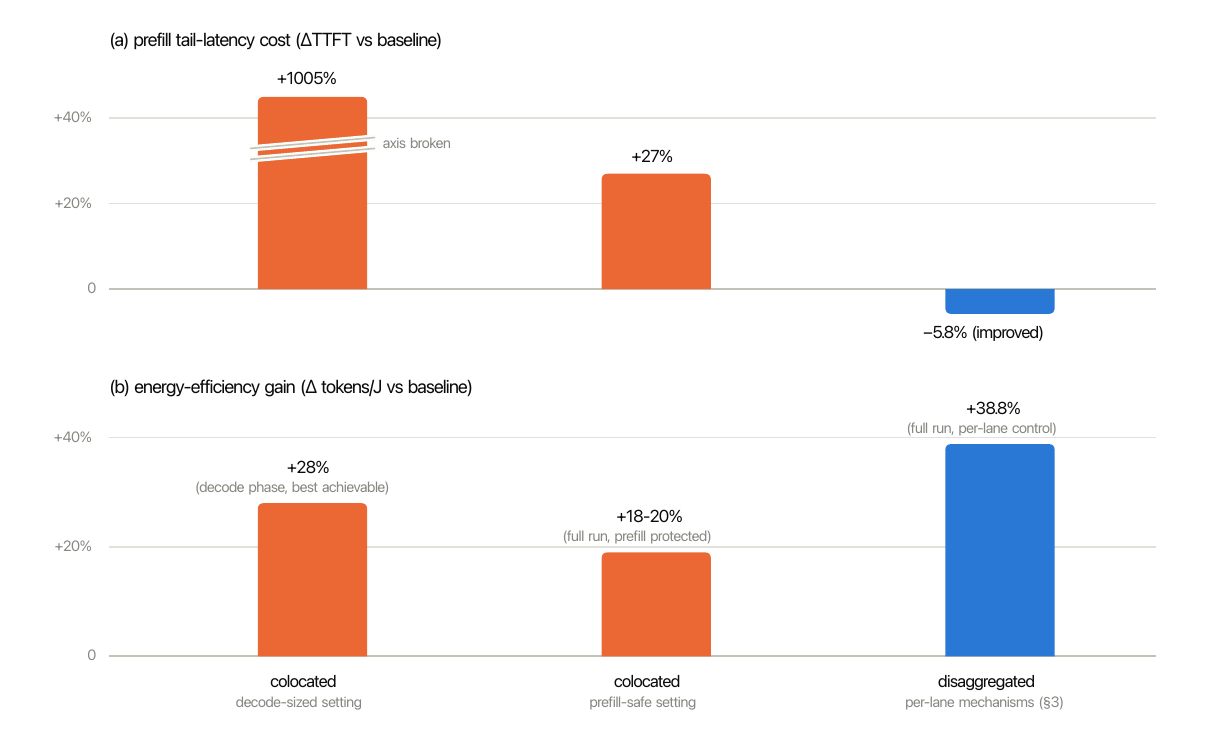}}
\caption{The colocated dilemma, and its structural resolution by disaggregation. Panels share the x axis: (a) prefill tail-latency cost and (b) energy-efficiency gain for the same three
configurations. A single colocated setting either collapses prefill latency (left, axis broken at +1005\%) or forgoes a large fraction of the decode-phase gain (middle); per-lane mechanisms on
disaggregated serving (right) obtain the largest efficiency gain while improving tail latency.}
\end{figure}

A second reason to target disaggregation directly: mechanism verdicts measured on colocated serving do not transfer. The combined cap-plus-clock-lock arm that gained +44\% tokens/J on the colocated
70B model retained only +4\% when the same model was redeployed PD-disaggregated. Any conclusion about actuation therefore had to be re-established on the disaggregated target itself, which is what
\S{}5.3 does.

\subsection{Why the power cap, and why MoE benefits most}\label{why-the-power-cap-and-why-moe-benefits-most}

The mechanistic asymmetry between the two in-band actuators is the foundation of the \S{}3.1 design. A locked clock stretches every instruction on the critical path, whether that path is
expert-dispatch communication (MoE) or long prefill compute (large dense models); a power cap is only an upper bound, under which the GPU schedules its own clocks, so most of the energy saving arrives
at near-zero latency cost. In the per-configuration sweep on the colocated 480B MoE model, the cap-only arm was Pareto-dominant outright: +30.8\% tokens/J with no inter-token-latency cost and full SLO
compliance.

The size of the cap-reclaimable gain is itself architecture-dependent (Figure A3). At comparable total scale, the cap-only arm recovered +26.0\ensuremath{\pm}6.4\% tokens/J on the 480B MoE but only
+5.7\ensuremath{\pm}8.0\% on the 405B dense model. The activity counters of A.1 explain the difference: dense models at frontier scale saturate the tensor pipes (prefill tensor activity 0.54-0.86) and
draw at or near the device power ceiling, leaving little for a cap to reclaim without cutting into demand; MoE models activate a fraction of their parameters per token (tensor activity 0.07-0.24), run
well below the ceiling, and pair that headroom with a flat, memory-bound decode profile that tolerates a binding cap. This is why the two MoE models of the main body yield +17.6\% to +38.8\%
(\S{}5.1), and it sharpens H1: the reclaimable slack, not just the setpoint, is a property of the deployed model.

\begin{figure}[tbp]
\centering
\pandocbounded{\includegraphics[keepaspectratio,alt={Why MoE benefits most. (a) At comparable total scale, cap-only control recovers roughly five times more efficiency on the MoE model than on the dense model. (b) The mechanism: dense models saturate compute and draw at the device ceiling; MoE models run far below saturation, leaving slack that a calibrated cap converts into savings.}]{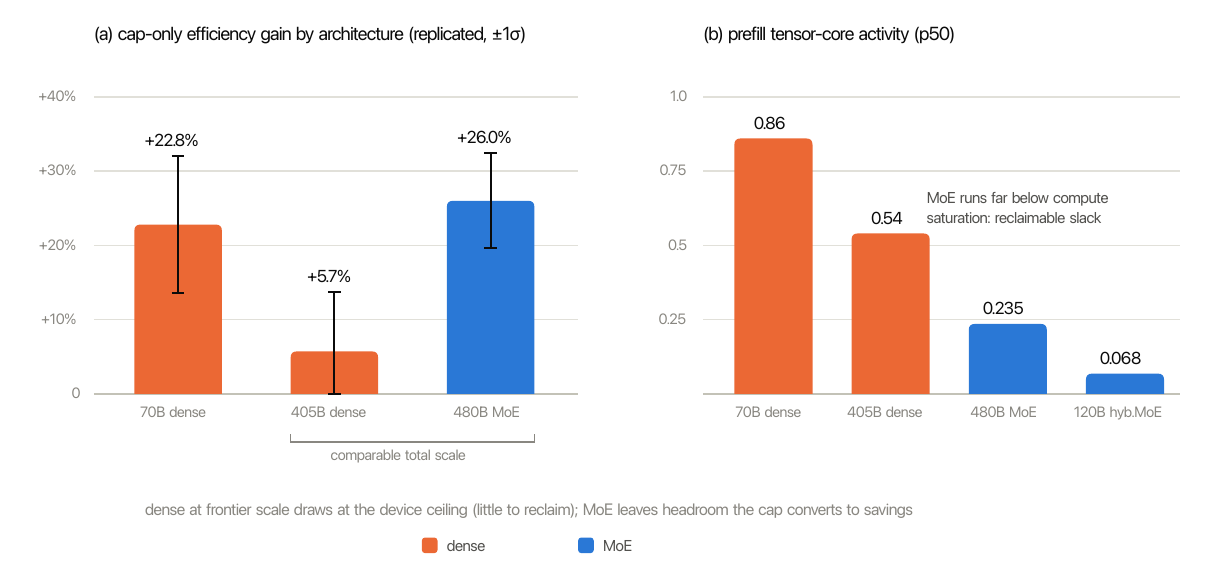}}
\caption{Why MoE benefits most. (a) At comparable total scale, cap-only control recovers roughly five times more efficiency on the MoE model than on the dense model. (b) The mechanism: dense models
saturate compute and draw at the device ceiling; MoE models run far below saturation, leaving slack that a calibrated cap converts into savings.}
\end{figure}

A methodological note from the same campaign: single-trial results on tail percentiles were unreliable enough to produce two conclusions that replication overturned, which set the replication
discipline used throughout \S{}5.

\subsection{Phase-decoupled actuation at run level}\label{phase-decoupled-actuation-at-run-level}

Figure A4 shows what the \S{}3.1 assignment does during steady disaggregated serving, per lane and per workload. The prefill lane keeps its bursty, compute-bound profile; the clock window lowers the
power peaks of each burst at the cost of a small stretch in burst duration, with the window floor bounding the latency impact by construction. The decode lane draws flat, memory-bound power; the cap
binds below the lane's natural draw for the entire run, cutting decode draw by 16\% (agentic) and 20\% (standard) while the GPU's self-DVFS preserves token cadence within the ITL-p99 SLO on both
workloads. The two mechanisms act simultaneously on different silicon, which is precisely what colocated serving cannot express.

\begin{figure}[tbp]
\centering
\pandocbounded{\includegraphics[keepaspectratio,alt={Lane-level power timelines under phase-decoupled control, both workloads (illustrative reconstruction from the PD serving A/B of 5.3; each lane normalized to its own natural draw, levels are measured per-lane means). The clock window shaves prefill burst peaks; the cap binds continuously through decode's flat draw.}]{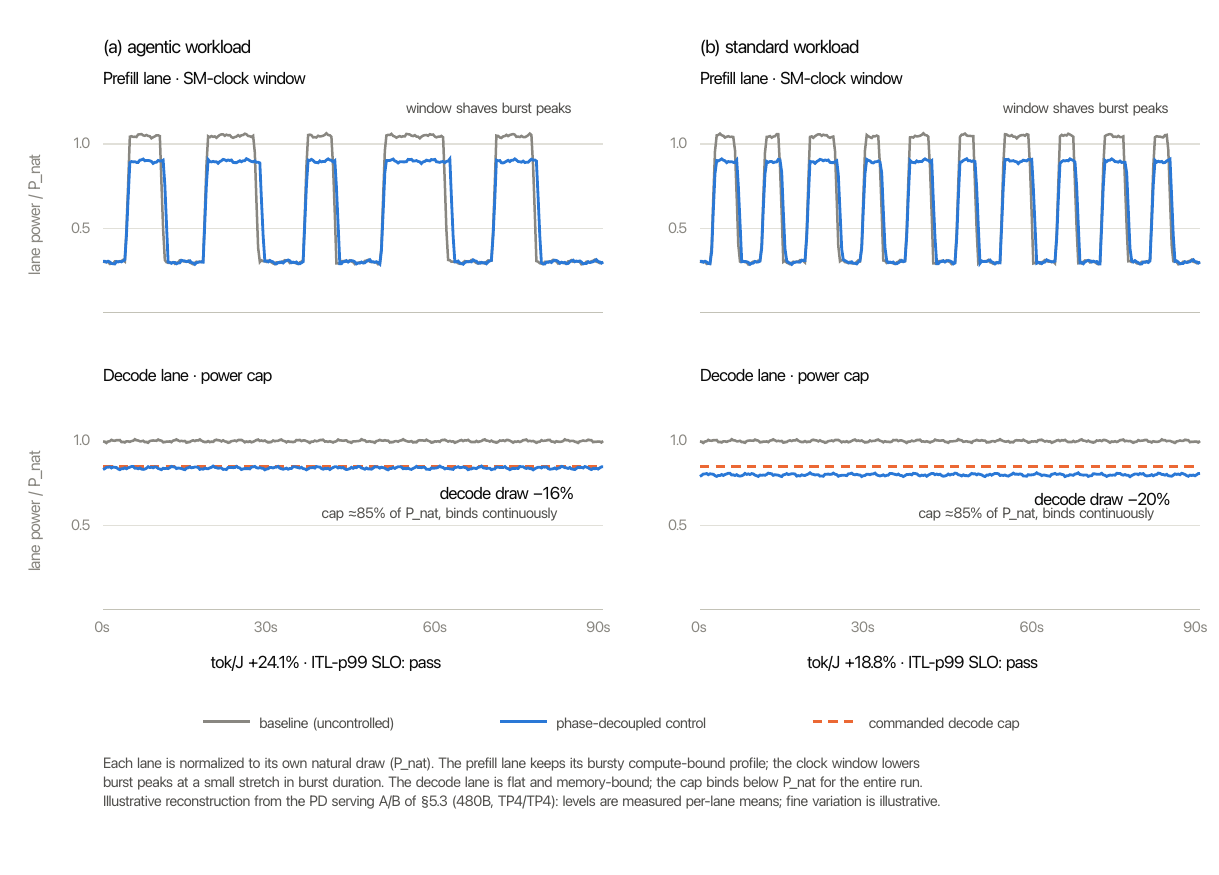}}
\caption{Lane-level power timelines under phase-decoupled control, both workloads (illustrative reconstruction from the PD serving A/B of \S{}5.3; each lane normalized to its own natural draw, levels
are measured per-lane means). The clock window shaves prefill burst peaks; the cap binds continuously through decode's flat draw.}
\end{figure}

The same protocol included a dry-run arm (control plane active, no hardware writes), which was indistinguishable from baseline on every metric; the control plane itself adds no measurable overhead.

\subsection{What the campaign established}\label{what-the-campaign-established}

\begin{enumerate}
\def\labelenumi{(\arabic{enumi})}
\tightlist
\item
  Mechanism and setpoint must be measured per deployment: every intermediate heuristic we formed was overturned either by replication or by a topology change (A.2). (2) The power cap was the only
  actuator whose latency cost remained bounded across all models, workloads, and topologies tested, which made it the decode-lane default that \S{}5.3 then confirmed head-to-head against static locks
  under SLO constraints. (3) Median metrics conceal actuation damage: a configuration can look lossless at the median while violating the tail SLO outright, which is why the acceptance criterion of
  \S{}3.3 and the evaluation lens of \S{}4 operate on tail percentiles. (4) Single-trial results on tail percentiles are unreliable; every campaign conclusion cited here is replicated.
\end{enumerate}


\begin{thebibliography}{19}\small
\bibitem{maxq} S.~Narayanaswamy, P.~Patel, I.~Karlin, U.~Gupta, V.~Saripalli, and Y.~Guo (NVIDIA). Datacenter energy optimized power profiles. arXiv:2510.03872, 2025.
\bibitem{polca} P.~Patel, E.~Choukse, C.~Zhang, \'I.~Goiri, B.~Warrier, N.~Mahalingam, and R.~Bianchini. Characterizing power management opportunities for LLMs in the cloud. In \emph{ASPLOS}, 2024.
\bibitem{splitwise} P.~Patel, E.~Choukse, C.~Zhang, A.~Shah, \'I.~Goiri, S.~Maleki, and R.~Bianchini. Splitwise: Efficient generative LLM inference using phase splitting. In \emph{ISCA}, 2024.
\bibitem{distserve} Y.~Zhong, S.~Liu, J.~Chen, J.~Hu, Y.~Zhu, X.~Liu, X.~Jin, and H.~Zhang. DistServe: Disaggregating prefill and decoding for goodput-optimized LLM serving. In \emph{OSDI}, 2024.
\bibitem{mooncake} R.~Qin et~al. Mooncake: A KVCache-centric disaggregated architecture for LLM serving. In \emph{USENIX FAST}, 2025.
\bibitem{voltana} J.~Yu, A.~Taneja, J.~Lin, and M.~Zhang. VoltanaLLM: Energy-efficient and SLO-aware disaggregated LLM serving via adaptive frequency control and state-space routing. In \emph{ISC High Performance}, 2026. arXiv:2509.04827.
\bibitem{dualscale} O.~Basit, Y.~Liu, Z.~J.~Kong, and Y.~C.~Hu. DualScale: Energy-efficient disaggregated LLM serving via phase-aware placement and DVFS. arXiv:2602.18755v3, 2026.
\bibitem{dynamollm} J.~Stojkovic, C.~Zhang, \'I.~Goiri, J.~Torrellas, and E.~Choukse. DynamoLLM: Designing LLM inference clusters for performance and energy efficiency. In \emph{HPCA}, 2025.
\bibitem{tapas} J.~Stojkovic, C.~Zhang, \'I.~Goiri, E.~Choukse, H.~Qiu, R.~Fonseca, and R.~Bianchini. TAPAS: Thermal- and power-aware scheduling for LLM inference in cloud platforms. In \emph{ASPLOS}, 2025.
\bibitem{rapid} Y.~Jiang, S.~Chowdhary, N.~Morris, R.~Jain, S.~Manne, and S.~Bayliss (AMD). RAPID: Power-aware dynamic reallocation for inference. arXiv:2601.12241, 2026.
\bibitem{pals} C.~Hankendi, R.~Shahout, M.~Yu, and A.~Coskun. PALS: Power-aware LLM serving for mixture-of-experts models. arXiv:2605.21427, 2026.
\bibitem{icpp-cap} Y.~Ma, S.~Subramaniyan, and X.~Wang. Power capping of GPU servers for machine learning inference optimization. In \emph{ICPP}, 2025.
\bibitem{greenllm} Q.~Liu, D.~Huang, M.~Zapater, and D.~Atienza. GreenLLM: SLO-aware dynamic frequency scaling for energy-efficient LLM serving. arXiv:2508.16449, 2025.
\bibitem{throttllem} A.~K.~Kakolyris, D.~Masouros, P.~Vavaroutsos, S.~Xydis, and D.~Soudris. throttLL'eM: Predictive GPU throttling for energy efficient LLM inference serving. In \emph{HPCA}, 2025. arXiv:2408.05235.
\bibitem{illusion} B.~Ma, A.~Afzal, J.~Eitzinger, and G.~Wellein. The illusion of power capping in LLM decode: A phase-aware energy characterisation across attention architectures. arXiv:2605.11999, 2026.
\bibitem{sharegpt} ShareGPT conversation corpus. \url{https://sharegpt.com}.
\bibitem{zeus} J.~You, J.-W.~Chung, and M.~Chowdhury. Zeus: Understanding and optimizing GPU energy consumption of DNN training. In \emph{NSDI}, 2023.
\bibitem{perseus} J.-W.~Chung, Y.~Gu, I.~Jang, L.~Meng, N.~Bansal, and M.~Chowdhury. Perseus: Reducing energy bloat in large model training. In \emph{SOSP}, 2024.
\bibitem{envpipe} S.~Choi, I.~Koo, J.~Ahn, M.~Jeon, and Y.~Kwon. EnvPipe: Performance-preserving DNN training framework for saving energy. In \emph{USENIX ATC}, 2023.
\end{thebibliography}
\end{document}